\documentclass{article}

   \usepackage[dvips]{graphicx}
   \graphicspath{{Figures/}}
	\usepackage{epsfig}

\usepackage{subcaption}
\usepackage{hyperref}

\usepackage[cmex10]{amsmath}
\usepackage{mdwmath}
\usepackage{mdwtab}
\usepackage{algorithm, algpseudocode}
\usepackage{amssymb,dsfont}

\usepackage{array}

\usepackage{url}

\usepackage{ulem}

\def \bY{\mathbf{Y}}
\def \bX{\mathbf{X}}
\def \bS{\mathbf{S}}
\def \bA{\mathbf{A}}

\def \b0{\mathbf{0}}
\def \bW{\mathbf{W}}

\def \bI{\mathbf{I}}
\newcommand{\bs}[1]{\mathbf{#1}}
\def \ik{^{(k)}}
\def \ikp{^{(k+1)}}

\def \conv_nGMCA{convolutive nGMCA}
\def \ana_nGMCA{analysis nGMCA}
\def \syn_nGMCA{synthesis nGMCA}
\def \ortho_nGMCA{orthonormal nGMCA}

\newcounter{prox_counter}
\newcommand{\labelprox}[1]{\refstepcounter{prox_counter}\arabic{prox_counter}\label{prox:#1}}

\title{NMF with Sparse Regularizations in Transformed Domains}

\author{J{\'e}r{\'e}my~Rapin\footnotemark[2]\ \footnotemark[3]
	\and J{\'e}r{\^o}me~Bobin\footnotemark[3]\ \footnotemark[4]
	\and Anthony~Larue\footnotemark[2]
	\and Jean-Luc~Starck\footnotemark[3]}

\begin{document}
\newcommand{\slugmaster}{%
\slugger{siims}{xxxx}{xx}{x}{x--x}}

\maketitle

\renewcommand{\thefootnote}{\fnsymbol{footnote}}
\footnotetext[2]{CEA, LIST, 91191 Gif-sur-Yvette Cedex, France.}
\footnotetext[3]{CEA, IRFU, Service d'Astrophysique, 91191 Gif-sur-Yvette Cedex, France.}
\footnotetext[4]{J.B. was supported by the French National Agency for Research (ANR) 11-ASTR-034-02-MultID.\\
Copyright \copyright ~by SIAM. Unauthorized reproduction of this article is 
prohibited.}
\renewcommand{\thefootnote}{\arabic{footnote}}

\begin{abstract}
Non-negative blind source separation (non-negative BSS), which is also referred to as non-negative matrix factorization (NMF), is a very active field in domains as different as astrophysics, audio processing or biomedical signal processing. In this context, the efficient retrieval of the sources requires the use of signal priors such as sparsity. If NMF has now been well studied with sparse constraints in the direct domain, only very few algorithms can encompass non-negativity together with sparsity in a transformed domain since simultaneously dealing with two priors in two different domains is challenging. In this article, we show how a sparse NMF algorithm coined non-negative generalized morphological component analysis (nGMCA) can be extended to impose non-negativity in the direct domain along with sparsity in a transformed domain, with both analysis and synthesis formulations. To our knowledge, this work presents the first comparison of analysis and synthesis priors ---as well as their reweighted versions--- in the context of blind source separation. Comparisons with state-of-the-art NMF algorithms on realistic data show the efficiency as well as the robustness of the proposed algorithms.
\end{abstract}

%
%

\pagestyle{myheadings}
\thispagestyle{plain}
\markboth{J.~RAPIN ET AL.}{NMF WITH SPARSE REGULARIZATIONS IN TRANSFORMED DOMAINS}

\section{Introduction}

Multichannel data are often encountered in scientific fields as different as geophysics, remote sensing, astrophysics or biomedical signal processing. In astrophysics for instance, the data are usually made of non-negative spectra measured at different locations. Each of these spectra is a mixture of several elementary source spectra which are characteristic of specific physical entities or sources. Recovering these sources is essential in order to identify the underlying components. Still, both the sources and the way they are mixed up together may be unknown. The aim of non-negative blind source separation (BSS) or non-negative matrix factorization (NMF) is to recover both the spectra and the mixtures.

\medskip
\subsection{Notations}

The following notations will be used throughout the article:
\begin{itemize}
\item $m$ is the number of measurements.
\item $n$ is the number of samples of the spectra/sources.
\item $r$ is the number of sources.
\item $\bX$ and all bold capital letters are matrices. The value of element $(i,j)$ is called $\bX_{i,j}$, row $i$ is $\bX_{i,\cdot}$ and column $j$ is $\bX_{\cdot,j}$. 
\item $\bY  \in \mathbb{R}^{m \times n}$ is the data matrix in which each row is a measurement.
\item $\bS \in \mathbb{R}^{r \times n}$ the unknown source matrix in which each row is a spectrum/source.
\item $\bA \in \mathbb{R}^{m \times r}$ the unknown mixing matrix which defines the contribution of each source to the measurements.
\item $\bs{Z} \in \mathbb{R}^{m \times n}$ is an unknown noise matrix accounting for instrumental noise and/or model imperfections.
\item $\| \bs{X} \|_p = \sqrt[p]{\sum_{i,j}|\bX_{i,j}|^p}$ (Frobenius norm for $p=2$).
\item $\odot$ and $\oslash$ are respectively the elementwise matrix multiplication (Hadamard product) and division.
\item $\bX \geq \bY$ means $\bX_{i,j} \geq \bY_{i,j},~\forall (i,j)$.
\item $i_\mathcal{C}$ is the characteristic function of the set $\mathcal{C}$. It is defined as follows:
\begin{align}
i_\mathcal{C}: x \mapsto &\begin{cases}
        0 \text{ if } x \in \mathcal{C},\\
        +\infty \text{ otherwise}.
    \end{cases}
\end{align}
\item $\bW\in\mathbb{R}^{p\times n}$ is a matrix transform from $\mathbb{R}^{n}$ to $\mathbb{R}^{p}$, $p$ being the dimension of the transformed domain. In this paper, we use transforms with $p\ge n$ such as orthonormal and redundant wavelets.
\item Signals in the transformed domain are marked with the subscript $w$. Since the sources in $\bS$ are row vectors, their transforms are provided by $\bS_w = \bS \bW^T$. When several subscripts are necessary, they are separated by the symbol $|$, for instance, the $j^\text{th}$ element of $x_w$ is denoted $x_{w|j}$.
\end{itemize}

\medskip
\subsection{Non-negative Matrix Factorization}
\label{sec:NMF}

In BSS, the instantaneous linear mixture model assumes that the measurements are linear mixtures of sources plus some noise. With the above notations, it can be written in the following matrix form:
\begin{equation}
\label{eq:BSSeq}
\bY=\bA \bS +\bs{Z}.
\end{equation}
Yet, it is common to have some prior information about the sources $\bS$ and the mixing matrix $\bA$. In the context of Non-negative Matrix Factorization (NMF, \cite{Paatero_94_Positivematrixfactorization, Lee_99_Learningpartsobjects}), the entries of both the mixture coefficients $\bA$ and the sources $\bS$ are assumed to be non-negative. This assumption arises naturally in many applications such as mass-spectrometry \cite{Dubroca_12_WeightedNMFhigh}, text mining \cite{Berry_05_EmailSurveillanceUsing}, clustering \cite{Kim_08_SparseNonnegativeMatrix}, audio processing \cite{Fevotte_09_Nonnegativematrixfactorization} or hyperspectral imaging \cite{Jia_09_ConstrainedNonnegativeMatrix}. Indeed, spectra are often measured as intensities (electromagnetic spectra for instance) or in terms of a whole number of elements (molecules in mass-spectrometry) which are necessarily non-negative. The mixture coefficients are usually function of the relative concentrations of the observed physical entities, which are necessarily non-negative as well. Under an i.i.d.\@ Gaussian noise assumption, the NMF problem is formulated as:
\begin{equation}
\label{eq:NMF}
\underset{\bA \ge \b0,~\bS \ge \b0}{\text{argmin}}~\| \bY-\bA\bS\|_2^2.
\end{equation}
The most well-known NMF algorithms are the multiplicative updates \cite{Lee_99_Learningpartsobjects, Lee_01_Algorithmsnonnegative} and the alternated least square (ALS, \cite{Paatero_94_Positivematrixfactorization}).

\medskip
\subsection{Additional Priors in NMF}
\label{sec:priors_NMF}

\begin{figure}[!t]
\centering
\includegraphics[width=3.35in]{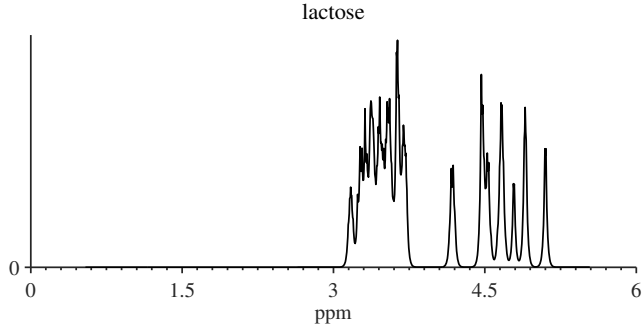}
\caption{A simulated Nuclear Magnetic Resonance (NMR) spectrum.}
\label{fig:lactose}
\end{figure}

Finding an optimal solution for \eqref{eq:NMF} is very complex since this problem is NP-Hard \cite{Vavasis_09_ComplexityNonnegativeMatrix}. Because there can be many local minima, and because the mixture model may be imperfect, adding other priors is usually beneficial. Indeed, they can help privileging minima with desired properties. A source such as the NMR (Nuclear Magnetic Resonance) spectrum displayed in figure \ref{fig:lactose} shows some structural features which have been exploited in several NMF algorithms.\medskip

\paragraph{Continuity of the sources} In \cite{Zdunek_07_BlindImageSeparation}, Zdunek \& Cichocki proposed to recover smooth and therefore continuous sources by adding a smoothing term to the problem in Equation~\eqref{eq:NMF}:
\begin{equation}
\label{eq:smoothNMF}
\underset{\bA \ge \b0,~\bS \ge \b0}{\text{argmin}}~\| \bY-\bA\bS\|_2^2+\alpha U_\delta(\bS),
\end{equation}
where $U_\delta$ is defined as the Green potential function:
\begin{equation}
\label{eq:green_potential}
U_\delta(\bS)=\delta\sum_{i=1}^r\sum_{j=2}^{n}\text{log}\left(\text{cosh}\left(\frac{\bS_{i,j}-\bS_{i,j-1}}{\delta}\right)\right).
\end{equation}
The minimization is then carried out using multiplicative updates. The regularization penalizes large differences between a sample and its neighbors and therefore tends to privilege smooth estimates of the sources. Still, tuning the parameters $\delta$ and $\alpha$ can be cumbersome and this regularization is not appropriate for spiky signals, such as the spectrum in figure \ref{fig:lactose}. A toolbox implementing this algorithm is available online\footnote{\url{http://www.bsp.brain.riken.jp/ICALAB/nmflab.html}}. Other types of smoothness regularizations have also been studied, such as the squared differences such as such as in \cite{Virtanen_07_MonauralSoundSource}.\medskip

\paragraph{Sparsity of the sources} In the wide sense, a sparse signal is a signal which concentrates its energy into only a few large non-zero coefficients, or can be well approximated in such a way. In many applications sources can be both sparse and non-negative, such as in mass or nuclear magnetic resonance (NMR) spectrometry for instance. The source in figure \ref{fig:lactose} is indeed sparse since it is composed of few large coefficients and many close to null coefficients. Algorithms have then been designed to recover such signals in NMF and stressed that sparsity could indeed help perform more relevant factorizations \cite{Kim_08_SparseNonnegativeMatrix, Eggert_04_Sparsecodingand, Hoyer_02_Nonnegativesparse}.\medskip

In \cite{Hoyer_02_Nonnegativesparse, Rapin_13_SparseandNon, Cichocki_07_HierarchicalALSAlgorithms}, sparsity is enforced by further constraining the $\ell_1$ norm of the sources as follows:
\begin{equation}
\label{eq:sparseNMF}
\underset{\bA\ge \b0,\bS\ge \b0}{\text{argmin}}~\|\bY-\bA\bS\|_2^2+\lambda \|\bS\|_1.
\end{equation}
In \cite{Hoyer_02_Nonnegativesparse}, Hoyer uses a gradient descent for $\bA$ and a multiplicative update procedure for $\bS$. In sparse HALS \cite{Cichocki_07_HierarchicalALSAlgorithms,Cichocki_09_NonnegativeMatrixand} columns of $\bA$ and rows of $\bS$ are updated one by one, which leads to simple and efficient updates. A recent accelerated implementation of HALS \cite{Gillis_12_AcceleratedMultiplicativeUpdates, Gillis_12_SparseandUnique} has been introduced. In this version of HALS, the parameter $\lambda$ is automatically handled in order to obtain a user-defined sparsity rate ---defined as the ratio of coefficients smaller than $10^{-6}$ times the largest one; it is available online\footnote{\url{https://sites.google.com/site/nicolasgillis/code}}. In \cite{Virtanen_07_MonauralSoundSource}, the authors also used this regularization along with a Kullback-Leibler divergence as data fidelity term. In non-negative Generalized Morphological Component Analysis (nGMCA, \cite{Rapin_13_SparseandNon}), the authors used proximal techniques in order to alternately and optimally estimate $\bA$ and $\bS$, with an automated parameter tuning. This work stands for the basis of the current paper and is further described in Section \S\ref{sec:transformed_nGMCA}. An implementation of the nGMCA algorithm is available online\footnote{\url{http://www.cosmostat.org/GMCALab.html}}.\medskip

Other forms of regularizations have also been explored to enforce the sparsity of the sources in the direct domain. In \cite{Kim_08_SparseNonnegativeMatrix} and \cite{Zdunek_07_Nonnegativematrixfactorization}, the authors have proposed the use of the penalization $\sum_{j=1}^n \|\bS_{\cdot ,j} \|_1^2$. This penalization tends to favor solutions where a single source dominates at each sample. In \cite{Hoyer_04_Nonnegativematrix}, Hoyer proposed to constrain a level of sparseness for each row of $\bS$. For a vector $x \in \mathbb{R}^n$, the sparseness value goes from 1 when $x$ is perfectly sparse ---only 1 active coefficient--- to 0 when all coefficients are active, with the same value. It is defined as follows:
\begin{equation}
\label{eq:HoyerSparseness}
\text{sparseness}(x)=\frac{\sqrt{n}-\frac{\|x\|_1}{\|x\|_2}}{\sqrt{n}-1}.
\end{equation}
\smallskip

If these algorithms are able to deal with spiky data, they do not take into account the continuity of source \ref{fig:lactose}. None of the aforementioned NMF algorithms is therefore able to use all the prior information about the sources. In order to better model this kind of signals, one can express sparsity in a different domain. Indeed, the sparsity of a signal depends on the basis or dictionary of waveforms in which it is expressed (see \cite{Elad_10_SparseandRedundant}). In a nutshell, bases that best capture the geometric structure of a signal will yield sparser representations of this signal. Sparsity-enforcing priors in transformed domains have been used with success to solve a very wide range of inverse problems (see \cite{Starck_10_SparseImageand} and references therein). In NMF however, few algorithms have been proposed to enforce sparsity in a different basis or dictionary, because of the difficulty in dealing with two different priors in two different domains. To our knowledge, only \cite{Jiang_12_Bregmaniterationalgorithm} has attempted to impose sparsity in a transformed domain. However this study is limited to imposing sparsity in an orthonormal basis. Furthermore, the estimation of $\bS$ is not carried out in an optimal way, which can clearly lead to sub-optimal performances as advocated in \cite{Rapin_13_SparseandNon}.

\medskip
\subsection{Contribution}

The aim of this article is to introduce a novel NMF algorithm enforcing the sparsity of the sources in a transformed domain, based on the nGMCA framework introduced in \cite{Rapin_13_SparseandNon} and the preliminary works in \cite{Rapin_13_SparseRedundantFormulations, Rapin_13_SparseRegularizationsand}. The nGMCA framework, which will be detailed in Section \S\ref{sec:nGMCA}, has been shown to yield efficient separation performance while providing an effective rule of thumb on how to set the sparsity parameter. However, it was so far limited to imposing sparsity in the direct domain.\medskip

In Section \S\ref{sec:transformed_nGMCA}, the nGMCA algorithm is extended to tackle sparsity in general transformed domains, either in orthonormal or redundant dictionary of waveforms. We further tackle the most common types of sparse priors in a transformed domain, namely the synthesis and the analysis formulations which, to the best of our knowledge, have never been compared in the context of NMF. Numerical comparisons with state-of-the-art NMF algorithms on noisy mixtures of simulated NMR spectra are carried out.\medskip

$\ell_1$-type priors yield estimation biases which can generally be neglected when the sources to be retrieved are very sparse ({\it e.g.} NMR and mass spectra). However for mildly sparse signals, such as natural images, the estimation bias related to the use of the $\ell_1$ norm hampers dramatically the performance. To alleviate this major issue, Section \S\ref{sec:reweighted-L1} presents an extension of the nGMCA algorithm which handles reweighted sparse priors \cite{Candes_07_EnhancingSparsityby}. Numerical experiments are carried out on natural images which show that the proposed algorithm yields enhanced separation performance.

\bigskip
\section{Presentation of nGMCA}
\label{sec:nGMCA}

\begin{algorithm}[!t]
\caption{: standard nGMCA}
\label{alg:nGMCA}
\begin{algorithmic}[1]
\Require $\bY$, $K$.
\State \textbf{initialize} $\bA^{(0)}$, $\bS^{(0)}$ and $\bs{\Lambda}^{(1)}$.
\For{$k\leftarrow 1,K$}
\State Normalize the columns of $\bA^{(k-1)}$.
\State $\bS^{(k)}\gets \underset{\bS\ge \b0}{\text{argmin}}~\frac{1}{2}\|\bY-\bA^{(k-1)}\bS\|_2^2 + \|\bs{\Lambda}^{(k)} \odot  \bS\|_1$.
\State $\bA^{(k)}\gets \underset{\bA\ge \b0}{\text{argmin}}~\frac{1}{2}\|\bY-\bA\bS^{(k)}\|_2^2$.
\State Select $\bs{\Lambda}^{(k+1)}\le\bs{\Lambda}^{(k)}$.
\EndFor
\State \textbf{return} $\bA^{(K)},~\bS^{(K)}$.
\end{algorithmic}
\end{algorithm}

\medskip
\subsection{Framework}

nGMCA \cite{Rapin_13_SparseandNon} was designed as an extension of GMCA \cite{Bobin_07_Sparsityandmorphological, Bobin_08_BlindSourceSeparation} to tackle NMF as formulated in problem \eqref{eq:sparseNMF}. The main steps of the algorithm are provided in \textbf{Algorithm \ref{alg:nGMCA}}. Each subproblem (lines 4 and 5) is solved at optimality, since it was observed through extensive experiments that optimally managing the priors enhances the separation performances (see \cite{Rapin_13_SparseandNon}). \medskip

An important feature of this algorithm is the decreasing $\ell_1$ regularization strategy. This kind of technique, inspired from simulated annealing, is also used in NMF with a decreasing $\ell_2$ or $\ell_{1,2}$ regularizations \cite{Zdunek_07_Nonnegativematrixfactorization, Cichocki_07_RegularizedAlternatingLeast}. In nGMCA, it consists in starting with large thresholds $\bs{\Lambda}$ so that the algorithm first estimates the mixing matrix from the entries of the sources which have the highest amplitude. Assuming that these large coefficients are likely to belong to only one source, they can indeed provide a descent separation. Next, the thresholds are decreased at each iteration in order to refine the solution. The final thresholds are set independently for each source at $\tau_\sigma^\infty \sigma^\text{grad}_i$, where $\sigma^\text{grad}_i$ is an online estimate of the noise level in the $i^\text{th}$ row of the gradient. This prevents most of the noise from altering the solution. The decrease of the thresholds is chosen to be linear, decreasing each threshold by a close to constant amount down to $\tau_\sigma^\infty \sigma^\text{grad}_i$.\footnote{in practice, the decrease is not exactly linear since the target value for the thresholds is updated at each new estimation of the noise level.}

\subsection{Convergence Analysis}
In \cite{Rapin_13_SparseandNon}, in order to allow the convergence of the algorithm, $\bs{\Lambda}$ is kept fixed during the last iterations. An additional constraint is also added to the update of $\bA$ (line 5 of \textbf{Algorithm \ref{alg:nGMCA}}) in order to thwart the scale indeterminacy: 

\begin{equation}
\text{5:}~~~~~\bA^{(k)}\gets ~ \underset{\bA\ge \b0,~\|\bA_{\cdot , j}\|^2_2 \le 1,\forall j}{\text{argmin}}~\frac{1}{2}\|\bY-\bA \bS^{(k)} \|_2^2. \label{eq:A_const_update}
\end{equation}

The full cost function of the problem it aims at solving is then:
\begin{equation}
F(\bA,\bS) = \|\bY-\bA\bS\|_2^2+ i_{\|\bA_{\cdot , j}\|^2_2 \le 1,\forall j}(\bA)+i_{\cdot\ge \bs{0}}(\bA) + \|\bs{\Lambda}\odot\bS\|_1+i_{\cdot\ge \bs{0}}(\bS).
\end{equation}
The minimization of this cost function is carried out here using block coordinate descent, i.e.\@ minimizing alternately the subproblems in $\bA$ and $\bS$. For this type of algorithm, Tseng \cite{Tseng_01_ConvergenceBlockCoordinate} provided convergence conditions when the cost function is not differentiable.\medskip

One of the conditions is that the levelsets of the cost function must be compact. This is the case for $F$ thanks to the additional constraint, indeed:
\begin{itemize}
\item the level sets of $F$ are closed, as the inverse images of closed sets of a continuous function.
\item they are also bounded in $\bS$ thanks to the $\ell_1$ regularization, and in $\bA$ thanks to the norm constraint on the columns of $\bA$.
\end{itemize}
Hence, as closed and bounded sets of a finite space, they are compact. Also, the data fidelity term $(\bA,\bS)\mapsto\|\bY-\bA\bS\|_2^2$ is differentiable on its domain, its domain is open, $F$ is continuous, and the subproblem in $\bA$ is pseudoconvex (since it is convex). This permits to use theorem 4.1(b) of \cite{Tseng_01_ConvergenceBlockCoordinate} which proves that the coordinate descent method with the cyclic rule converges to a stationary point of $F$. More details are provided in Example 6.4 of \cite{Tseng_01_ConvergenceBlockCoordinate} which is closely related to the present settings.

\medskip
\subsection{Optimization Procedures - Proximal Splitting}

In order to solve the subproblems at lines 4 and 5, nGMCA makes use of a proximal splitting method, namely the forward-backward algorithm \cite{Combettes_05_Signalrecoveryby}. This algorithm is able to tackle problems which take the following form:
\begin{equation}
\underset{x}{\text{argmin}}~f(x)+g(x),\label{eq:f+g}
\end{equation}
where $f$ is a convex and differentiable function; and $g$ is a non-differentiable proper convex and lower semi-continuous function. It alternates between local minimization of $f$ and of $g$ until convergence, which occurs under mild assumptions. To do so, the algorithm requires the gradient of $f$ and the proximal operator of $g$. The proximal operator of a proper convex lower semi-continuous function $g$ is defined as: 
\begin{equation}
\text{prox}_g(x)=\underset{y}{\text{argmin}}~\frac{1}{2}\|y-x\|_2^2+g(y).
\end{equation}
Proximal operators take a closed-form expression for a wide range of functions $g$; Table \ref{tab:operators_1} further features the expression of the proximal operators of interest in this paper. \medskip

\begin{table}[!t]
\footnotesize
\centering
\begin{tabular}{|c|c|c|}
\hline
  \# & Function & Proximal operator\\
\hline
\hline
 \labelprox{pos} & $i_{\cdot \ge 0}(x)$ & $[x]_+=\text{max}(x,~0)$\\
\hline
 \labelprox{L1} & $ \|\lambda \odot x\|_1$ & $\text{Soft}_\lambda(x)= \text{sign}(x)\odot [~|x|-\lambda ~ ]_+$\\
\hline
\labelprox{L1pos} & $\|\lambda \odot x\|_1+i_{\cdot \ge 0}(x)$ & $[ \text{Soft}_\lambda(x) ]_+$\\
\hline
\labelprox{constr_pos} & $i_{.~\ge 0,~\|.\|_2^2\le 1}(x)$ & $[x]_+/\text{max}\left(\|[x]_+\|_2,~1 \right)$\\
\hline
\end{tabular}
\caption{Some functions and their proximal operators \label{tab:operators_1} (with $x$ a column vector).}
\end{table}

In the case of the update of $\bS$,  $f(\bS)=\frac{1}{2}\|\bY-\bA\bS\|_2^2$ is the data fidelity term; and $g(\bS)=i_{\cdot\ge \b0}(\bS) +\lambda\|\bS\|_1$ accounts for the sparse prior and the non-negativity constraint. Its proximal operator is called non-negative soft-thresholding and is given by proximal \#\ref{prox:L1pos} of table \ref{tab:operators_1}. The subproblem in $\bA$ is solved similarly, with proximal operator \#\ref{prox:pos}, or with proximal operator \#\ref{prox:constr_pos} in the norm-constrained case of problem \eqref{eq:A_const_update} (derivation of this operator is provided in appendix \ref{app:norm_constrained_prox}).

\bigskip
\section{Transformed non-negative GMCA}
\label{sec:transformed_nGMCA}

In the context of BSS, sparsity has been shown to provide more diversity or contrast between the sources which greatly help improving their separation \cite{Zibulevsky_99_BlindSourceSeparation, Li_03_Sparserepresentationand, Bobin_07_Sparsityandmorphological}. Enforcing sparsity in a transformed domain makes possible the separation of sources with complex geometrical structures (see \cite{Bobin_07_Sparsityandmorphological}). The aim of this section is to explore extensions of the nGMCA algorithm to tackle NMF problems with sparsity imposed in a transformed domain. It has to be noticed that, in contrast to current sparse constraints in the direct domain, dealing with two priors ---non-negativity and sparsity--- expressed in different domains is challenging. In the sequel, we will particularly focus on two formulations of sparse regularization in a transformed domain: synthesis and analysis. To the best of our knowledge, this is the first study of an analysis regularization in the context of source separation.

\medskip
\subsection{Synthesis and Analysis Formulations}

In a transformed domain, sparsity can be enforced in two different ways, namely synthesis and analysis formulations.\medskip

With $\bW \in \mathbb{R}^{p \times n}$, minimizing a function $f:\mathbb{R}^n \rightarrow\mathbb{R}$ with a synthesis regularization is expressed in the following way:
\begin{equation}
\underset{x_w\in \mathbb{R}^p}{\text{argmin}}~f(\bW^Tx_w)+ \lambda\|x_w\|_1.
\end{equation}
In this formulation, the unknown of the minimization is not directly the signal but sparse coefficients in the transformed space $\hat{x}_w\in \mathbb{R}^p$. Denoting $\bs{D} = \bW^T$, the aim is to reconstruct the sought signal $\hat{x}$ as a sparse linear combination of columns of $\bs{D}$ ---i.e.\@ using as few columns of $\bs{D}$ as possible---:
\vspace{-0.4cm}
\begin{equation}
\hat{x}= \bW^T \hat{x}_w = \sum_{j=1}^p \hat{x}_{w|j}D_{.,j}.
\vspace{-0.3cm}
\end{equation}
$\bs{D}$ is consequently called a dictionary and its columns are called atoms. This is a generative model, hence the name ``synthesis": one builds the sought-after signal using bricks of the signal space, the atoms.\medskip

Still, it is essential to notice that the fact that $\hat{x}_w$ is sparse does not mean that $\bW\hat{x}=\bW\bW^T\hat{x}_w$ is necessarily sparse, since usually $\bW\bW^T\neq \bs{I}$. In practice, this means that the synthesis formulation finds a solution which has a sparse representation in the transformed domain, but not that the transform of the solution is sparse. In the analysis formulation, one therefore directly carries out the minimization in the direct/signal domain in order to find a solution which is sparse when multiplied by $\bW$. This is expressed as follows:
\vspace{-0.1cm}
\begin{equation}
\underset{x\in \mathbb{R}^n}{\text{argmin}}~f(x)+ \lambda\|  \bW x\|_1. 
\vspace{-0.15cm}
\end{equation}

The term $\lambda \| \bW x\|_1$ penalizes correlations between $x$ and the atoms/columns of $\bs{D} = \bW^T$. In other words, while in the synthesis formulation the signal was expressed as a sum of a limited number of atoms, the aim in the analysis formulation is to obtain a signal which is strongly correlated with only a few atoms of $\bs{D}$ and which is weakly correlated with the other atoms.\medskip

When $\bW$ is orthonormal, the synthesis and the analysis formulations are strictly equivalent \cite{Selesnick_09_SignalRestorationwith}. Indeed, the change of variables $x_w = \bW x$ lets us obtain one formulation from the other, since $\bW$ is invertible in this case. Still, one may want to use redundant dictionaries, with $p>n$, since they were shown to enhance signal recoveries \cite{Coifman_95_Translationinvariantde}. The advantage of redundant dictionaries comes from the larger number of atoms available to sparsely represent signals. In redundant wavelets for instance, this allows translation invariance \cite{Coifman_95_Translationinvariantde}. 
With such redundant transforms, analysis and synthesis formulations were then shown to have very different behaviors. Indeed, since $p\neq n$, the minimization spaces in the synthesis and analysis formulations are necessarily different. In the field of inverse problems, the latter was observed to be more flexible and better adapted to natural signals, which cannot generally be synthesized from a few atoms \cite{Elad_07_AnalysisVersusSynthesis, Selesnick_09_SignalRestorationwith, Nam_13_CosparseAnalysisModel}.

\begin{table}[!t]
\footnotesize
\centering
\begin{tabular}{|c|c|c|}
\hline
  \# & Function & Proximal operator\\
\hline
\hline
 \labelprox{wavePosOrth} & $i_{(\bs{R}^T \cdot) \ge 0}(x)$ & $\bs{R}[\bs{R}^T x]_+$\\
\hline
 \labelprox{wavePos} & $i_{(\bW^T \cdot) \ge 0}(x)$ & $x+\bW[- \bW^T x]_+$\\
\hline
 \labelprox{analysis} & $\|\lambda \odot \bW x \|_1$& $x-\bW^T(\underset{|u_{w|i}| \le\lambda_i,\forall i}{\text{argmin}}\|x-\bW^T u_w\|_2^2)$\\
\hline
 \labelprox{Linf} & $i_{|\cdot_i|\le\lambda_i,\forall i}(x)$& $P^\infty_\lambda(x)=x-\text{Soft}_\lambda(x)$\\
\hline
\end{tabular}
\caption{Some functions and their proximal operators\label{tab:operators_2}, with $x$ a column vector, $\bs{R}\in \mathbb{R}^{n\times n}$ an orthonormal transform, and $\bW\in \mathbb{R}^{p\times n}$ with $p\ge n$ a transform such that $\bW^T \bW =\bs{I}$ (tight frame)}
\end{table}

\medskip
\subsection{Synthesis nGMCA}

In order to extend the nGMCA algorithm to use a sparse synthesis regularization, lines 4 and 5 of \textbf{Algorithm \ref{alg:nGMCA}} have to be updated as follows:

\noindent\hrulefill
\vspace{-0.2cm}
\begin{align}
\text{4:}~~~~~&\bS_w^{(k)}\gets ~ \underset{ \bS_w \bW \ge \b0}{\text{argmin}}~\frac{1}{2}\|\bY-\bA^{(k-1)}\bS_w\bW \|_2^2  + \| \bs{\Lambda}^{(k)} \odot  \bS_w\|_1.\\
\text{5:}~~~~~&\bA^{(k)}\gets ~ \underset{\bA\ge \b0}{\text{argmin}}~\frac{1}{2}\|\bY-\bA\bS^{(k)}_w\bW \|_2^2. \label{eq:syn_update}
\vspace{-0.4cm}
\end{align}
\noindent\hrulefill\medskip

Since the variable is now $\bS_w$ and not $\bS$, one must also provide $\bS_w^{(0)}$ instead of $\bW^{(0)}$ at the beginning and set $\bS^{(K)}=\bS_w^{(K)}\bW$ at the end.\medskip

Although the update of $\bA$ is only slightly modified, the additional transform $\bW$ provides a new difficulty for the update of $\bS$. Indeed, there is no convenient way to compute the proximal operator of $g(\bS_w)=\|\bs{\Lambda}\odot\bS_w\|_1+i_{\cdot\bW\ge \b0}(\bS_w)$ and it is therefore not practical to use the forward-backward algorithm in this case.\medskip

Instead, one can apply the generalized forward-backward algorithm (GFB) \cite{Raguet_13_GeneralizedForwardBackward}, which considers $g$ as the sum of two convex lower semi-continuous functions $g_1(\bS_w)= \|\bs{\Lambda}\odot\bS_w\|_1$ and $g_2(\bS_w)= i_{\cdot\bW\ge \b0}(\bS_w)$.
This algorithm also requires their proximal operators. The proximal operator of $g_1$ is  proximal \#\ref{prox:L1} from table \ref{tab:operators_1} but the proximal operator of $g_2$ is not always analytic. If $\bW$ is orthonormal, the operator admits a closed form and is provided as proximal \#\ref{prox:wavePosOrth} from table \ref{tab:operators_2}. In the experiments, such settings are used in the algorithm coined  ``\ortho_nGMCA" with an orthonormal wavelet transform.

As stated previously, redundant transforms were shown to improve reconstructions. In the particular case of transforms $\bW \in \mathbb{R}^{p \times n}$ with $p>n$ and such that $\bW^T\bW=\bI$ (tight frames), the proximal operator is still analytic. It is provided as proximal \#\ref{prox:wavePos} in table \ref{tab:operators_2} and is derived in appendix \ref{app:synProx}. This proximal operator generalizes proximal operator \#\ref{prox:wavePosOrth} which could only handle orthonormal, and therefore non-redundant, transforms. The pseudo-code of the synthesis update step is provided in \textbf{Algorithm \ref{alg:synthesis_update}} of appendix \ref{app:syn_update}. 
In the experiments, this algorithm is then coined ``\syn_nGMCA" and is used with a redundant wavelet transform.\medskip

For even more general transforms $\bW$, the proximal operator of $g_2$ does not admit a closed-form expression anymore. In this case, it would be more convenient to use an algorithm such as Chambolle-Pock \cite{Chambolle_10_firstorderprimal}, as will be the case for the analysis formulation below.

\medskip
\subsection{Analysis nGMCA}

In order to adapt the nGMCA algorithm to use a sparse analysis regularization in a transformed domain, line 4 of \textbf{Algorithm \ref{alg:nGMCA}} has to be updated as follows:

\noindent\hrulefill
\begin{align}
\text{4:}~~~~~&\bS^{(k)}\gets ~ \underset{\bS\ge \b0}{\text{argmin}}~\frac{1}{2}\|\bY-\bA^{(k-1)}\bS\|_2^2+ \| \bs{\Lambda}^{(k)} \odot  (\bS\bW^T)\|_1.\label{eq:ana_update}
\end{align}
\noindent\hrulefill\medskip

In \cite{Rapin_13_SparseRedundantFormulations} and \cite{Rapin_13_SparseRegularizationsand}, the authors use redundant wavelet transforms and carry the minimization in the same way as in the synthesis formulation, but with proximal \#\ref{prox:pos} and \#\ref{prox:analysis} instead of proximal \#\ref{prox:wavePos} and \#\ref{prox:L1}. Yet, proximal \#\ref{prox:analysis}, which is derived in appendix \ref{app:anaProx}, is no more analytic and needs to be computed through a subroutine (using the FB algorithm with proximal \#\ref{prox:Linf} for instance). It therefore requires intensive computations. Several other optimization schemes are however possible in order to avoid the need of a subroutine. In Appendix \ref{app:analysis_chambolle} we show how to use the Chambolle-Pock algorithm \cite{Chambolle_10_firstorderprimal} for the resolution of \eqref{eq:ana_update} with any transform $\bW$ and provide the pseudo-code (\textbf{Algorithm \ref{alg:analysis_update}}). This approach is the one used for the ``\ana_nGMCA" in the experiments, with a redundant wavelet transform.

\subsection{Numerical Complexity}
Each iteration of the updates of $\bS$ involves a matrix multiplication, a forward and a backward transform, and linear steps such as the thresholding, the projections on the non-negative orthant and linear combinations. The matrix multiplication involves an $r\times r$ matrix and $\bS$, and has therefore a complexity of $\mathcal{O}(r^2n)$. With $\phi(n)$ the complexity of the transform on a signal of length $n$, the complexity of the transformations is in $\mathcal{O}(r\phi(n))$, which yields an overall complexity of the order of $\mathcal{O}(r\phi(n)+r^2n)$.

In this article, we mostly use redundant wavelet transforms which have a complexity of $\phi(n)= n~\text{log}_2(n)$ \cite{Coifman_95_Translationinvariantde}. The numerical complexity is therefore linearithmic in the number of samples $n$ of each source (i.e.\@ in $n~\text{log}_2(n)$) because of the transform, and quadratic in the number of sources $r$ because of the matrix multiplication.\medskip

Note that, in any case, the number of observations $m$ does not appear at all in the complexity of the update of $\bS$. The number of observations indeed only appears ---linearly--- in the update of $\bA$, which has complexity of $\mathcal{O}(r^2m)$ because of the matrix multiplication.

\subsection{Experiments}

\begin{figure}[!t]
\centering
\includegraphics[width=3.35in]{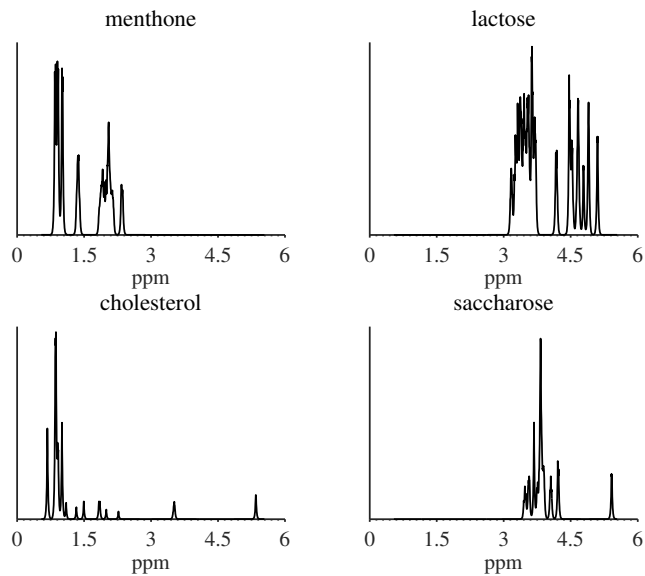}
\caption{NMR spectra of four natural compounds.}
\label{fig:NMRspectra}
\end{figure}

\begin{figure}[!t]
\centering
\includegraphics[width=3.35in]{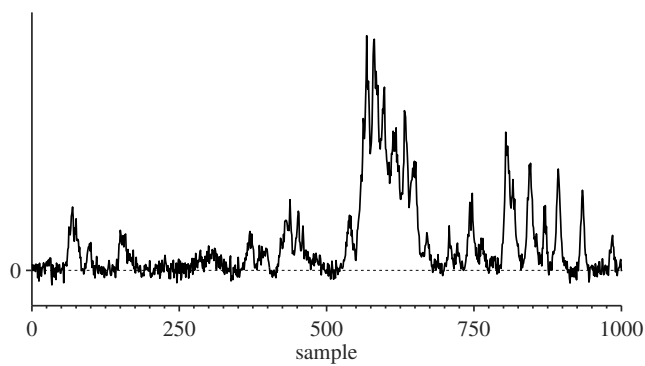}
\caption{Example of noisy mixture ($r=15$, $15$dB)}
\label{fig:appli_mixture}
\end{figure}

\subsubsection{Experimental settings and Evaluation}
\label{sec:settings_eval}
In physical applications, a molecule can be identified by its specific Nuclear Magnetic Resonance (NMR) spectrum. In this section, we evaluate the algorithms on simulated data made of mixtures of realistic NMR spectra:
\begin{enumerate}
\item the mixing matrix $\bA$ is drawn as the absolute value of an i.i.d.\@ Gaussian random matrix.
\item the spectra in $\bS$ come from the Spectral Database for Organic Compounds, SDBS\footnote{\url{http://riodb01.ibase.aist.go.jp/sdbs/cgi-bin/cre_index.cgi}}. In order to account for the acquisition imperfections, the spikes were convoluted with a Laplacian kernel of 4-samples width at half-maximum (width-4 Laplacian kernel). Some of the obtained sources are shown in figure \ref{fig:NMRspectra}.
\item the data matrix $\bY$ is generated as $\bY=\bA\bS+\bs{Z}$ where $\bs{Z}$ is a Gaussian noise matrix. In the experiments, it gathers $m=32$ measurements of $n=1024$ samples. An example of mixture is provided in figure \ref{fig:appli_mixture}.
\end{enumerate}

The sources are naturally non-negative and already significantly sparse in the direct domain. They can however benefit from wavelet-sparsity since they are close to piecewise polynomial. In the following, we consider $\bW$ to be 3-level wavelets (Symmlets-4). The orthonormal version uses Wavelab\footnote{\url{http://www-stat.stanford.edu/~wavelab/}} and the redundant ones a homemade reimplementation of the Rice wavelet toolbox\footnote{\url{http://dsp.rice.edu/software/rice-wavelet-toolbox}} (see Section \S\ref{sec:software}). Both satisfy $\bW^T \bW=\bI$ (tight frames) after adequate renormalization when necessary. With these settings, the mean sparseness (as defined in equation \eqref{eq:HoyerSparseness}) is 0.78 in the direct domain and 0.89 in the orthonormal wavelet domain. All the nGMCA-based algorithms are left running for 300 iterations.  $\tau^\infty_\sigma=1$ for nGMCA and the orthonormal version and 2 for the analysis and synthesis versions.\medskip

These algorithms are compared with smooth NMF \cite{Zdunek_07_BlindImageSeparation} and the accelerated sparse HALS \cite{Gillis_12_SparseandUnique}, which are described in Section \S\ref{sec:priors_NMF}, and are left running until convergence. In each simulation, sparse HALS is provided with a sparsity level which is pre-computed from the actual reference sources. Since there is no straightforward way to tune the smoothness parameter $\alpha$ in problem \eqref{eq:smoothNMF} for smooth NMF, we keep the best reconstruction out of 12 different value of the parameter $\alpha$ (logarithmically distributed between $10^{-5}$ and $1$). In this sense, both these algorithms are provided with close to optimal parameters while the nGMCA based algorithms are not.\medskip

In order to evaluate the outputs of BSS algorithms, Vincent et al.\@ \cite{Vincent_06_Performancemeasurementin} have proposed to decompose the estimated sources into the sum of several contributions through a sequence of projections:
\begin{equation*}
s^\text{est}=s_\text{target}+s_\text{interf}+s_\text{noise}+s_\text{artifacts},
\end{equation*}
with the following interpretations for the terms: 
\begin{itemize}
\item $s_\text{target}$ is the projection of $s^\text{est}$ on the target (ground-truth) source. In other words, it is the one part of this decomposition which corresponds to what needs to be recovered. The other ones are residues.
\item $s_\text{interf}$ accounts for interferences due to other sources.
\item $s_\text{noise}$ is the part of the reconstruction which is due to noise.
\item $s_\text{artifacts}$ stands for the remaining artifacts which are neither due to interferences nor noise.
\end{itemize}
\medskip

Using this decomposition, the authors designed scale-invariant SNR-type energy ratios in order to evaluate the detrimental influence of interferences (through an SIR criterion), of noise (SNR), and of artifacts (SAR) in the results. They also designed a more global criterion named Source to Distortion Ratio (SDR):
\begin{equation}
\text{SDR}(s^\text{est})=10~\text{log}_{10}\left(\frac{\|s_\text{target}\|_2^2}{\|s_\text{interf}+s_\text{noise}+s_\text{artifacts}\|_2^2}\right).\label{eq:SDR}
\end{equation}
All these criteria increase for better recoveries. The latter criterion is intensively used below in order to evaluate the algorithms since it takes into account all the aspects of a good reconstruction (low interferences, noise and artifacts). In the experiments it is computed through Monte-Carlo simulations. The number of simulations is provided in each figure's caption.\medskip

\subsubsection{Convolutive model}
In this section, the sources are modeled as sparse spike trains convoluted with a Laplacian kernel. The sources can therefore be decomposed as a non-negative linear combination of non-negative atoms. In this case, the problem of the update of $\bS$ can be greatly simplified since non-negativity and sparsity can be expressed in the same domain:
\begin{equation}
\bS^{(k)}\gets ~ \underset{ \bS_w}{\text{argmin}}~\frac{1}{2}\|\bY-\bA^{(k-1)}\bS_w\bW \|_2^2  + \| \bs{\Lambda}^{(k)} \odot  \bS_w\|_1 +i_{\cdot\ge \b0}(\bS_w).
\end{equation}
The advantage of this formulation is that function $g(\bS_w)= \|\bs{\Lambda} \odot\bS_w\|_1+i_{\cdot \ge \b0}(\bS_w)$ has an analytic proximal operator and the minimization can be carried with the FB algorithm such as in standard nGMCA. In the sequel, we use a convolution matrix $\bW$, such that for a column vector $x$, $\bW x = f \ast x$, with $f$ a convolutive kernel. This version of nGMCA is coined ``\conv_nGMCA". The convolutive kernel of convolutive nGMCA is set to a width-4 Laplacian kernel. When the sources are indeed composed of non-negative spikes convoluted with such kernel, this approach should lead the best separation performances; it will therefore play the role of a reference method in this section.\\

\begin{figure}[!t]
\centering
\includegraphics[width=3.35in]{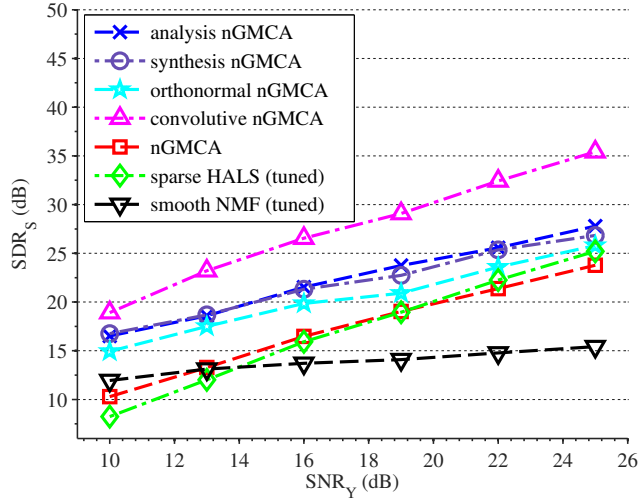}
\caption{Reconstruction SDR (SDR$_S$) with respect to the noise level in dB (simulated NMR spectra, $r=12$, $m=32$, average of $36$ simulations)}
\label{fig:noise_nmr}
\end{figure}

\begin{figure}[!t]
\centering
\includegraphics[width=3.35in]{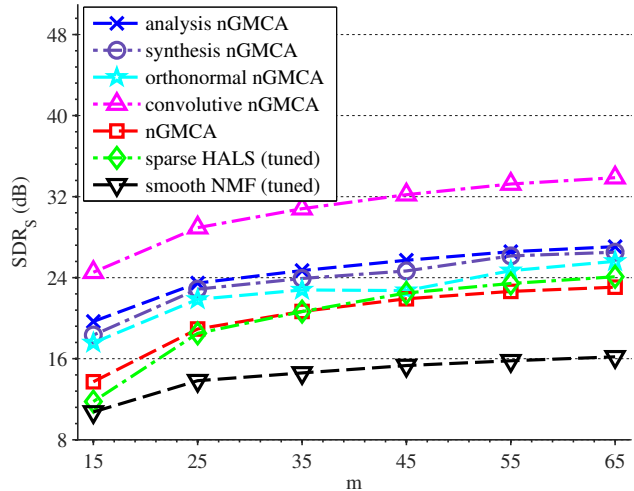}
\caption{Reconstruction SDR (SDR$_S$) with respect to the number of measurements $m$ (simulated NMR spectra, $r=12$, SNR$_Y=20$dB, average of $36$ simulations)}
\label{fig:m_nmr}
\end{figure}

\begin{figure}[!t]
\centering
\includegraphics[width=3.35in]{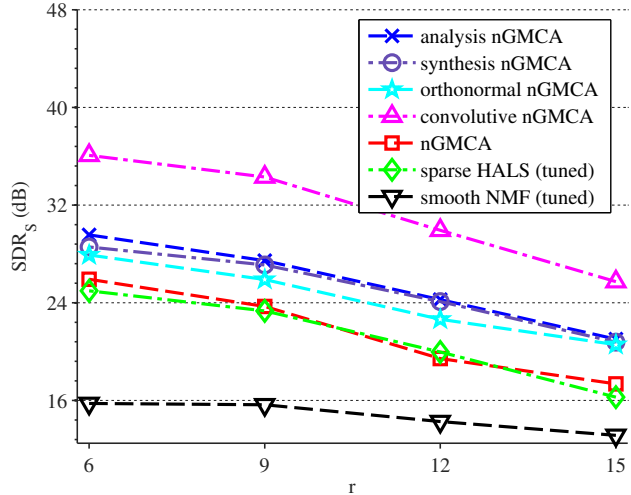}
\caption{Reconstruction SDR (SDR$_S$) with respect to the number of sources $r$ (simulated NMR spectra, SNR$_Y=20$dB, $m=32$, average of $36$ simulations)}
\label{fig:r_nmr}
\end{figure}

\subsubsection{Comparison of the formulations}
Figures \ref{fig:noise_nmr}, \ref{fig:m_nmr} and \ref{fig:r_nmr} show the performances of the nGMCA-based algorithms compared to other methods, respectively with varying noise levels, number of measurements $m$ and number of sources $r$. The results generally indicate a common trend: smooth NMF does not perform as well as sparsity-enforcing methods (nGMCA, sparse HALS) for this type of sources as it is not well adapted to cope with singularities, such as the peaks which compose the sources. Wavelets are better suited for this goal \cite{Mallat_92_Singularitydetectionand} and imposing sparsity in an orthonormal wavelets consequently yields increased performance of about a couple of dB. As expected, redundant wavelets, which are translation-invariant, whether used in the synthesis formulation or in the analysis formulation, yield enhanced separation results.\\
As emphasized previously, the sources to be retrieved can be exactly modeled as a non-negative and sparse mixture of atoms of $\bW$ ({\it i.e.} convolutive model). The nGMCA algorithm based on the convolutive model performs best in this case.\\
It is important to notice that, in the nGMCA-based algorithms, the regularization parameter is set automatically. This is not the case for the other algorithms which often turn to be more difficult to parameterize. In particular, it is not clear how the parameter of smooth NMF must be tuned. In the experiments, smooth NMF are optimally tuned so that they lead to the best reconstructions based on the knowledge of the sought after sources. In sparse HALS, the parameter is set to the sparsity level of the true sources. Obviously, this would be unavailable in practice.\medskip

\begin{figure}[!t]
\centering
\includegraphics[width=3.35in]{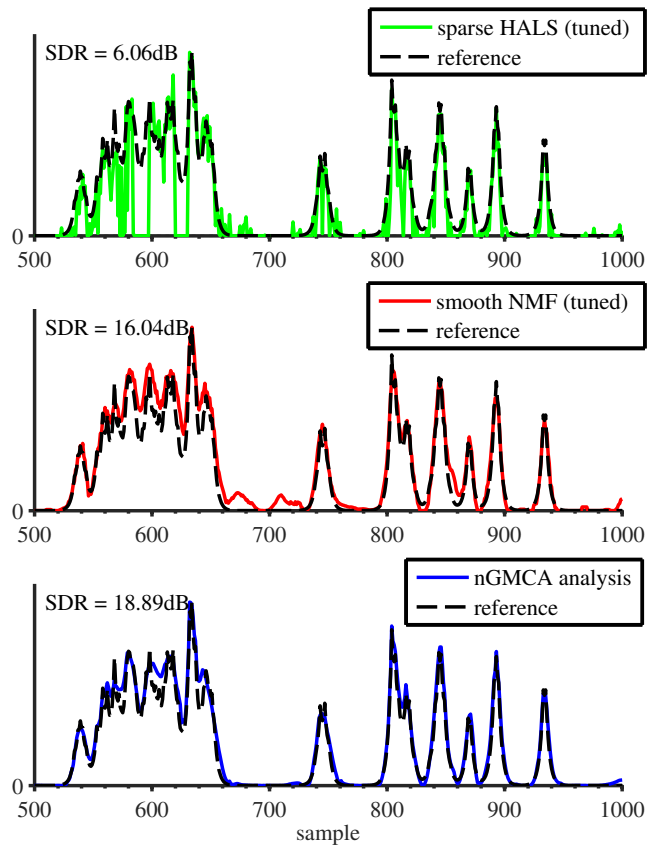}
\caption{Examples of reconstruction ($r=12$, $m=32$, SNR$_Y=10$dB)}
\label{fig:appli_reconstruction_example}
\end{figure}

As expected, \conv_nGMCA yields the best results as generative convolutive model is exact in this particular case. This information is however usually unknown or is even not reliable in practice; this will be discussed later on. Smooth NMF is not able to recover the sources properly: while it is well adapted to recover smooth geometric structures, it is not well suited to recover singularities such as the peaks of the sources. nGMCA and sparse HALS perform slightly better since the sources are mildly sparse in the direct domain. Still, as stated above, the sources are sparser in the orthonormal wavelet domain (sparseness of 0.89 compared to 0.78 in the direct domain), which explains the couple of additional dB provided by \ortho_nGMCA. The translation invariance of redundant wavelets is also significantly beneficial as shown with \ana_nGMCA and \syn_nGMCA which perform from 2 to 4 dB even better, with a neat advantage for the analysis formulation. This corroborates the results in \cite{Elad_07_AnalysisVersusSynthesis, Selesnick_09_SignalRestorationwith} for natural signals: the sources can not be exactly synthesized from a few coefficients in $\bW$. In this case, the analysis-based regularization is generally more robust and yields better reconstruction results.\medskip

Figure \ref{fig:appli_reconstruction_example} shows an example of reconstruction for \ana_nGMCA, sparse HALS and smooth NMF and confirms our analysis. Indeed, the smoothing effect of smooth NMF is clearly illustrated with the peaks being smoothed out. It also highlights that the sparse regularization in the direct domain is not completely adapted since it does not capture the smooth structure; it essentially tends to bias the sources towards $0$. Wavelets have the ability to sparsely represent local singularities ---peaks--- as well as smooth and large-scale geometric structures. Enforcing sparsity in the wavelet domain allows for a better recovery of both the sharp peaks and the large-scale structures of the signals. \medskip

\begin{figure}[!t]
\centering
\includegraphics[width=3.35in]{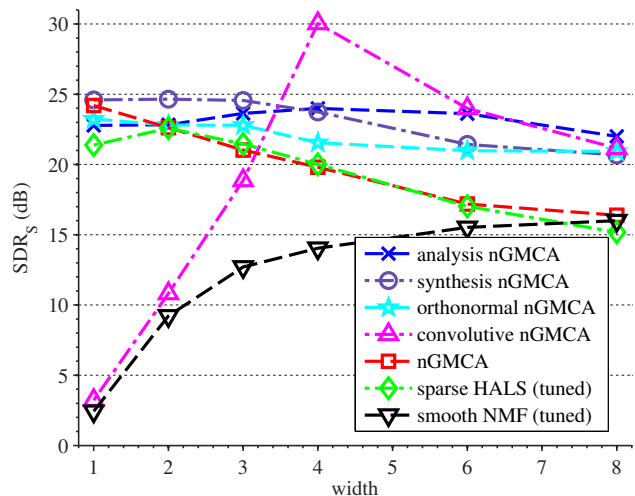}
\caption{Reconstruction SDR (SDR$_S$) with respect to the width of the peaks (simulated NMR spectra, SNR$_Y=20$dB,  $r=12$, average of $36$ simulations)}
\label{fig:width_nmr}
\end{figure}

\begin{figure}[!t]
\centering
\includegraphics[width=3.35in]{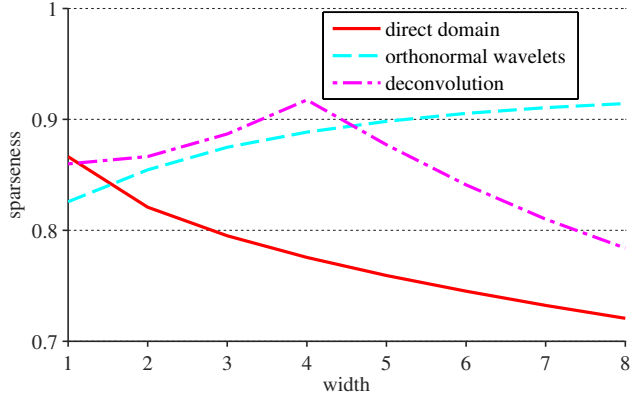}
\caption{Mean sparseness of the NMR sources in the different domains.}
\label{fig:sparseness_width}
\end{figure}

It is also very instructive to have a look at the importance of the peaks width. In figure \ref{fig:width_nmr}, the width at half-maximum of the peaks is modified compared to the previous cases where it was fixed to $4$. This highly impacts the model, since the smaller the width, the spikier the data. Unsurprisingly, smooth NMF does not perform well at recovery spiky data. Conversely, standard nGMCA performs best for extremely spiky spectra. Its performances with respect to \syn_nGMCA and \ana_nGMCA decrease when the peaks' width increases. Finally and most interestingly, the convolutive kernel of \conv_nGMCA is left unchanged in this experiment. It is then clearly visible that the performances of the algorithm decrease very quickly when the kernel is no more perfectly tuned to the data (which is the case only at width $4$). One should definitely establish a parallel between these observations and figure \ref{fig:sparseness_width} which provides the mean sparseness of the sources in the direct domain, in the wavelet domain, and after an inversion of the width-4 Laplacian kernel, for several kernel widths. When sparseness in the direct domain is a valid assumption, standard nGMCA performs better than wavelet sparse nGMCA versions. Conversely, the versions of nGMCA in the wavelet domain outperform standard nGMCA when the data are sparser in the wavelet domain. The \conv_nGMCA algorithm appears to be much more sensitive to the accuracy of the model.

\medskip

\subsubsection{Summary}
Through these experiments, we have seen that the additional structural information provided by the domain transform is helpful for the reconstruction of sources with complex geometrical structures which cannot be correctly tackled by smooth NMF or standard sparse NMF methods. This is especially true in difficult settings such as in the low measurement case or when the number of sources is large. Compared to the  $\ell_2$-based smoothness regularization, sparsity in a well-chosen domain ---here the wavelet domain--- also has the advantage of preserving the peaks while correctly reconstructing large-scale smooth components. Lastly, the automatic and straightforward strategy used to handle the sparsity parameters in the nGMCA-based algorithms is efficient for a wide range of settings.\\
To the best of our knowledge, these experiments also present the first comparison of synthesis and analysis-based regularizations in the scope of blind source separation and NMF. As already emphasized in the context of linear inverse problems \cite{Elad_07_AnalysisVersusSynthesis, Selesnick_09_SignalRestorationwith}, analysis-based problems are generally more robust to model mismatch as, in contrast to the synthesis-based regularization, does not assume that the signals to be retrieved are decomposed into a few non-zero entries in $\bW$.

\bigskip
\section{Reweighted-L1 in BSS}
\label{sec:reweighted-L1}

In this section, we observe the limitations of the $\ell_1$ regularization and explain how it is possible to bypass it. This approach is then validated on noisy mixtures of images.

\begin{figure}[!t]
\centering
\includegraphics[width=3.35in]{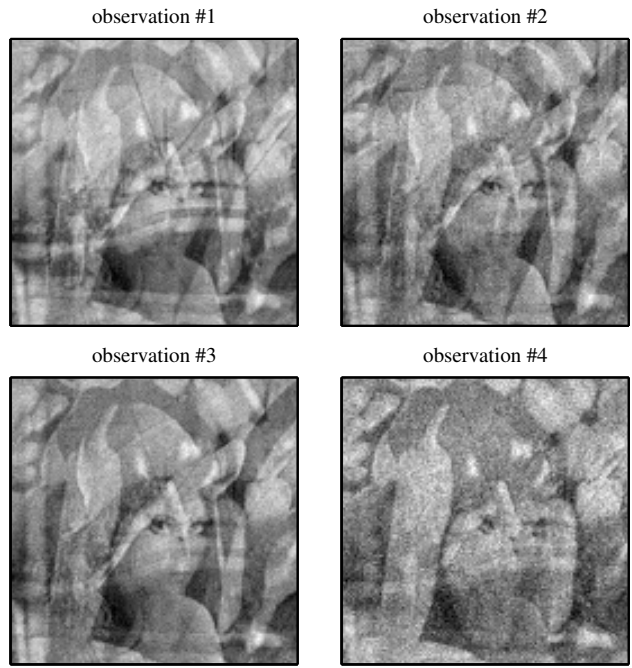}
\caption{$4$ noisy mixtures out of $32$ ($25$dB).}
\label{fig:4_mixtures_2D}
\end{figure}

\begin{figure}[!t]
\centering
\includegraphics[width=3.35in]{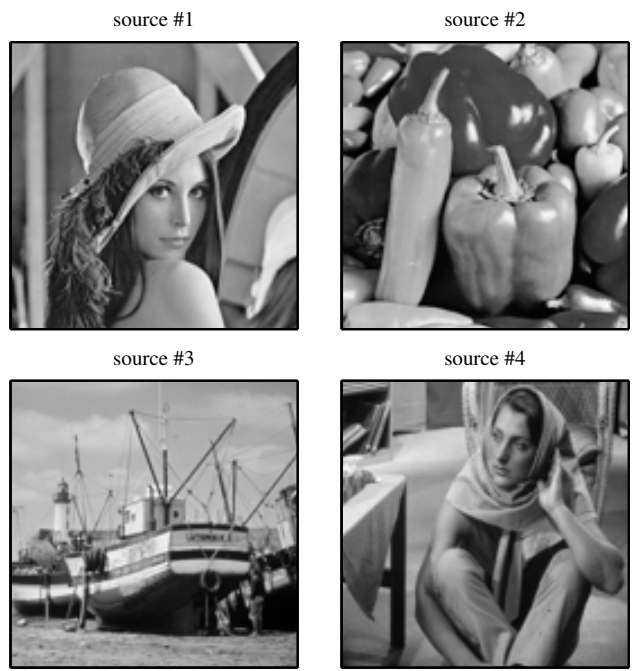}
\caption{Ground truth 2D sources ($128\times128$ pixels).}
\label{fig:4_sources_2D}
\end{figure}

\medskip
\subsection{Bias and Interferences induced by the $\ell_1$ Regularization}

In article \cite{Rapin_13_SparseandNon}, the authors discussed the pros and cons of the customarily used $\ell_1$ regularization in the context of non-negative BSS. More specifically, while this regularization can achieve superior separation performances, it is well known to produce a bias which can hamper the separation performances.\medskip 

\begin{figure}[!t]
	\hspace{0.1\linewidth}
	\begin{subfigure}[b]{0.35\linewidth}
		\includegraphics[width=\linewidth]{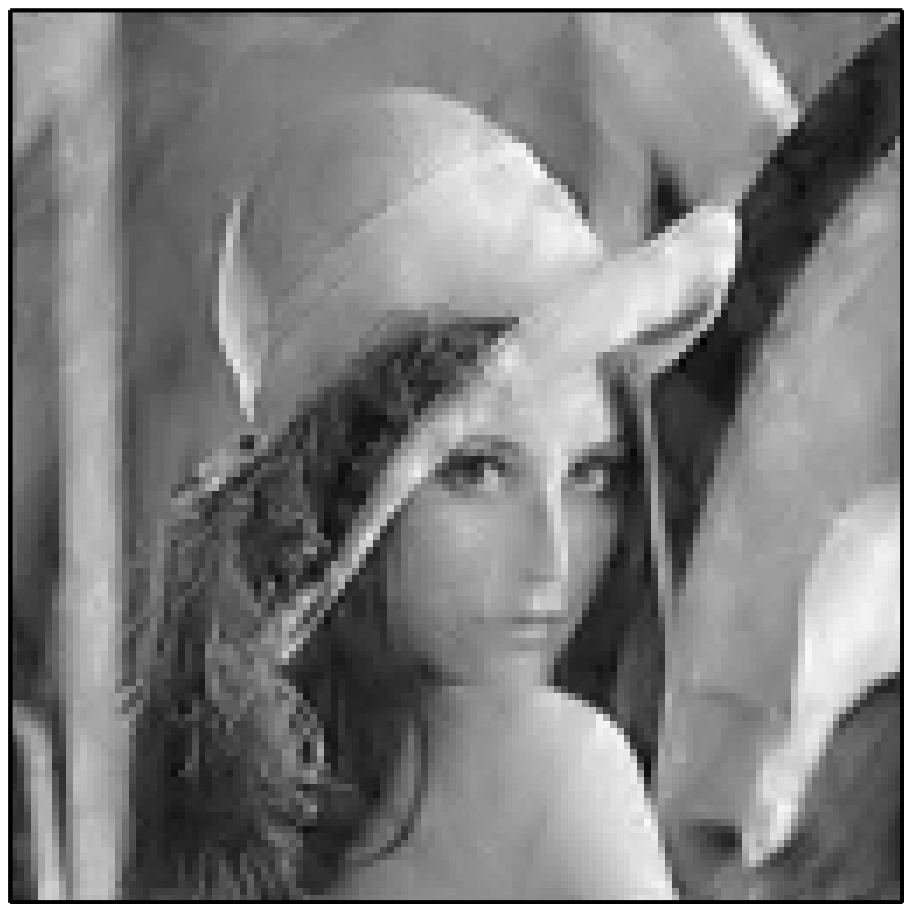}
	\caption{regular-$\ell_1$ regularization\\ (SDR$ = 21.77$dB for this source).}
		\label{fig:lena_inversion_regular}
    \end{subfigure}%
    \hspace{0.1\linewidth}
	\begin{subfigure}[b]{0.35\linewidth}
		\includegraphics[width=\linewidth]{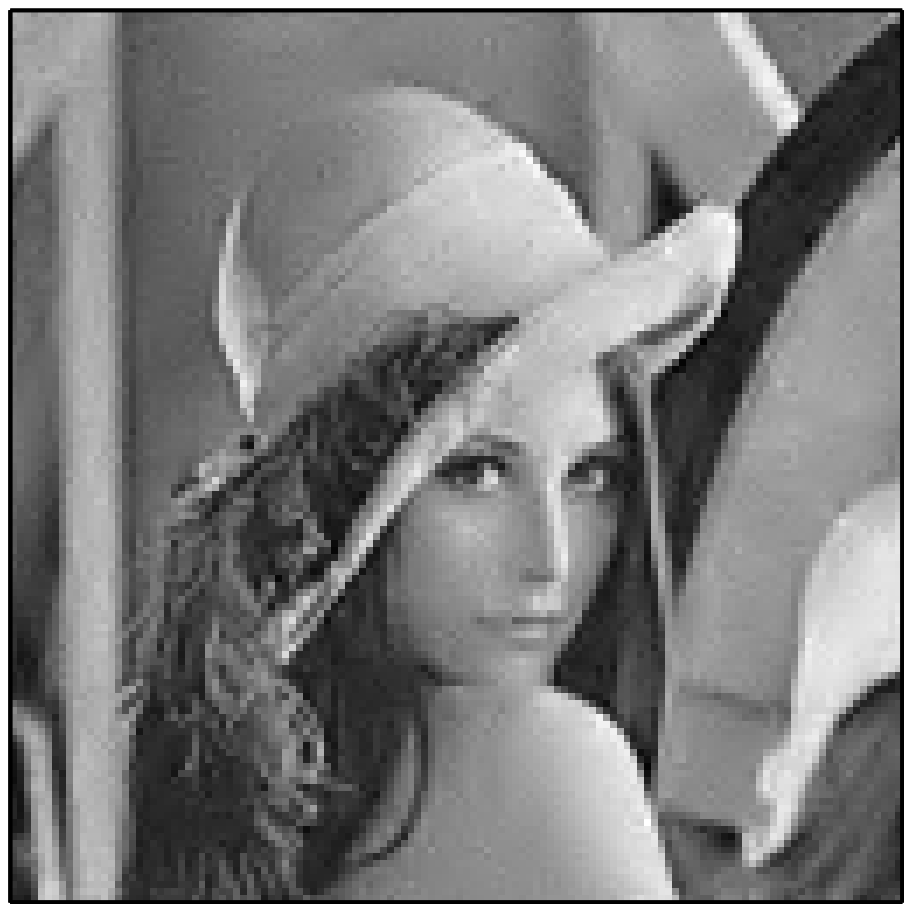}
		\caption{reweighted-$\ell_1$ regularization (SDR$ = 26.62$dB for this source).}
		\label{fig:lena_inversion_reweighted}
	\end{subfigure}
	\caption{Examples of reconstruction for Lena\\(SNR$_Y = 25$dB, $\tau = 0.05$).\vspace{-0.3cm}}
	\label{fig:house_inversion}
\end{figure}

If the bias induced by the $\ell_1$ regularization can be neglected for very sparse sources ({\it i.e.} the number of active entries in the sources is very low), it is hardly the case for more complex data which are rather modeled as approximately sparse signals. In that case, all the entries of the sources are non-zero but only a few take a significant amplitude.\\
For instance, natural images are good examples of approximately sparse signals in the wavelet domain. In this setting, the $\ell_1$ regularization may not be as effective.\\
Figure \ref{fig:4_mixtures_2D} shows 4 out of 32 noisy mixtures of the 4 sources displayed in figure \ref{fig:4_sources_2D} (namely: Lena, peppers, a boat and Barbara), the mixture coefficients being the absolute value of a Gaussian matrix. Figure \ref{fig:lena_inversion_regular} shows the recovery of Lena after solving problem \eqref{eq:S_analysis} with the ground truth mixing matrix $\bA$.\\
One can observe strong interferences: some parts of the peppers and Barbara's head being visible on the picture.\medskip

In the framework of standard inverse problems, the $\ell_1$ regularization is customarily substituted with the $\ell_0$ regularization which generally produce a lower estimation bias. In the context of sparse NMF \cite{Rapin_13_SparseandNon}, the authors observed that an $\ell_0$ penalization do not exhibit similar interferences, but does not perform as well as the $\ell_1$ regularization. This is especially the case in difficult separation instances ({\it e.g.} when the number of sources to be retrieved is large). The correct separation of approximately sparse signals ---such as natural images--- therefore requires a special treatment by taking the best of the $\ell_1$ and  $\ell_0$ penalizations.

\medskip
\subsection{Balancing between $\ell_1$ to $\ell_0$}

In the context of regularized linear inverse problems, several techniques have been designed so as to overcome the bias produced by the $\ell_1$ penalization. In \cite{Gao_97_Waveshrinkwith}, the authors have introduced a mixture of the soft- and hard-thresholding named firm-thresholding. In \cite{Voronin_13_newiterativefirm}, the authors have proposed to iteratively alter the penalization from soft- to firm thresholding ; this strategy makes profit of the convexity of the $\ell_1$ regularization at the beginning of the algorithm. This allows to uniquely get a first estimate of the solution which is assumed to be close to the signal to be retrieved.\medskip

In this article, we opt for an alternative called reweighted-$\ell_1$ \cite{Candes_07_EnhancingSparsityby}. This method consists in choosing the coefficients of the regularization matrix $\bs{\Lambda}$ according to the amplitude of the corresponding coefficients of the transform of $\bS$. The entries of the transform of $\bS$ with the largest amplitudes will be less penalized than the small amplitude entries. This is usually done by iteratively solving a sequence of weighted $\ell_1$-regularized problem. At each step of the sequence, the weights $\bs{\Lambda}$ are re-estimated from the current estimate of the sources $\bS$. The main advantage of this method is that, although it globally requires solving a non-convex problem, each step remains convex.\medskip

Several reweighting strategies have been proposed \cite{Candes_07_EnhancingSparsityby}. In the current paper, the weight matrix $\bs{\Lambda}$ will be updated as follows:

\begin{equation}
\bs{\Lambda}^\text{rew.}_\tau = \bs{\Lambda}_\tau \oslash \left(\bs{1} + \left(\bS_{\text{inv}|w} \oslash \bs{\Sigma}\right)^2\right), \label{eq:Lambda_rew}
\end{equation}
where the square is applied elementwise, and $\bS_{w|\text{inv}}$ is the least-square estimate of the sources using the current estimate of $\bA$:
\begin{equation}
\bS_{w|\text{inv}} = (\bA^T\bA)^{-1}\bA^T \bY \bW^T. \label{eq:S_inv_w}
\end{equation}
The matrix $\bs{\Sigma}$ contains the noise standard deviation of the sources in the transform domain. In practice, it is evaluated by using a robust empirical estimator of the noise standard deviation based on the MAD (Median Absolute Deviation) of $\bS_{w|\text{inv}}$.\medskip

Applied to the same mixture of natural images, the reweighted $\ell_1$ regularization is used to estimate the sources assuming $\bA$ is known; one of the sources is displayed in Figure \ref{fig:lena_inversion_reweighted}. For the sake of evaluation, we use the same criteria as defined in Section~\ref{sec:settings_eval}. These quantities are provided in Table \ref{tab:reweighting} in the following settings: i) no regularization is applied (non-negative least-square estimate), ii) standard $\ell_1$ regularization and iii) reweighted $\ell_1$ regularization. In comparison to a simple least-square estimate, the use of the $\ell_1$ regularization obviously helps removing noise contamination ---high SNR--- but at the cost of more interferences and artifacts (lower SIR and SAR). As featured in the last column, the reweighted-$\ell_1$ penalization provides a good balance between denoising and separation efficiency, since the SNR slightly decreases but the SIR greatly increases. 

\begin{table}[!t]
\footnotesize
\centering
\begin{tabular}{|c|c|c|c|}
\hline
  criterion & no regularization & regular-$\ell_1$ & reweighted-$\ell_1$\\
\hline
\hline
SDR$_S$ & 23.5 & 20.9 & 25.8 \\
\hline
SIR$_S$ & 64.9 & 25.7 & 35.4 \\
\hline
SNR$_S$ & 23.5 & 37.0 & 30.6 \\
\hline
SAR$_S$ & 51.1 & 22.9 & 28.5 \\
\hline
\end{tabular}
\caption{Reconstruction criteria on an inversion with regular and reweighted $\ell_1$ (SNR$_Y=25$dB, regularization parameter at $\lambda = 0.05$).}
\label{tab:reweighting}
\end{table}

\medskip
\subsection{Experiments}

\begin{figure}[!t]
\centering
\includegraphics[width=3.35in]{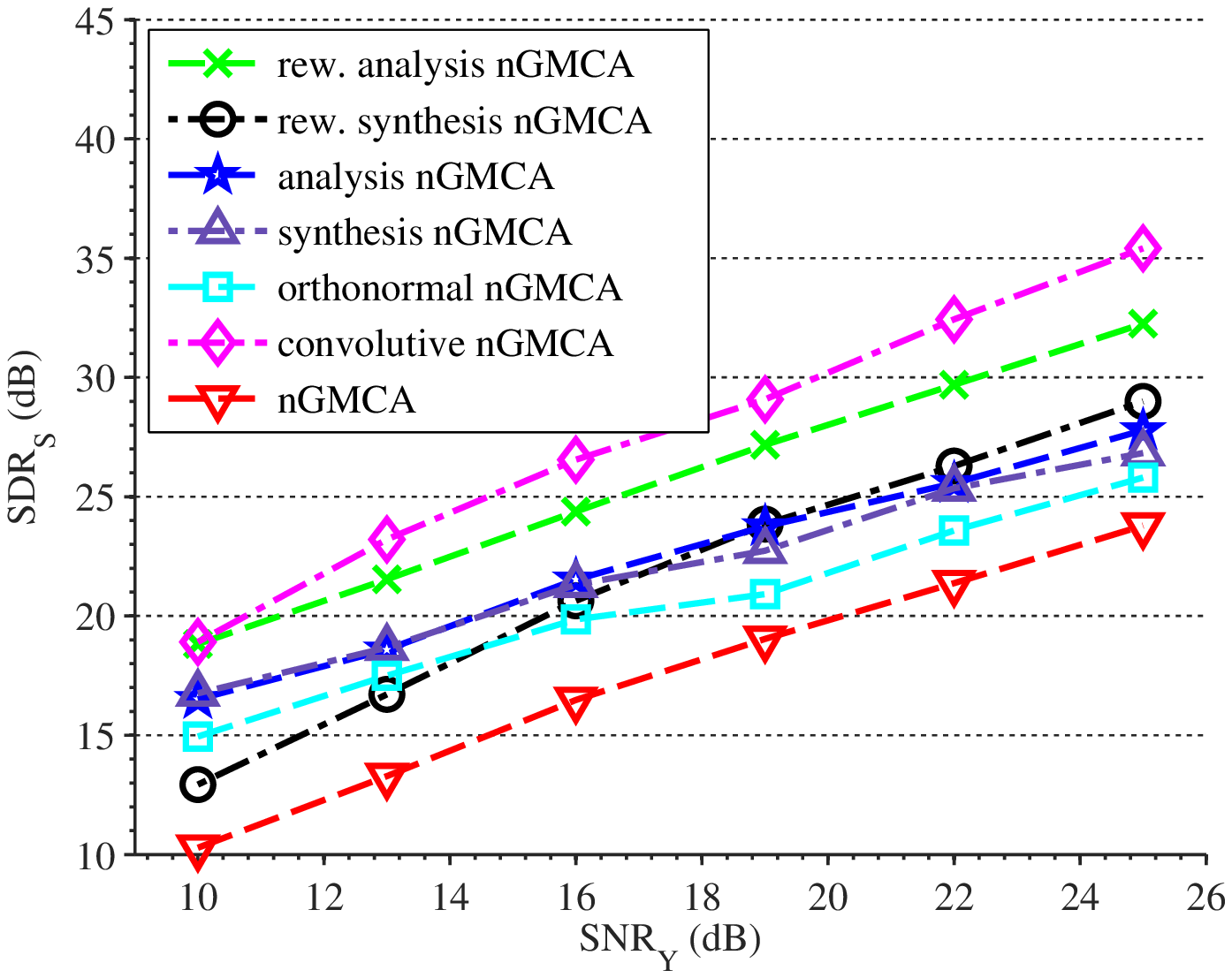}
\caption{Reconstruction SDR (SDR$_S$) with respect to the noise level in dB (simulated NMR spectra, $r=12$, $m=32$, average of $36$ simulations)}
\label{fig:noise_nmr_rew}
\end{figure}

\begin{figure}[!t]
\centering
\includegraphics[width=3.35in]{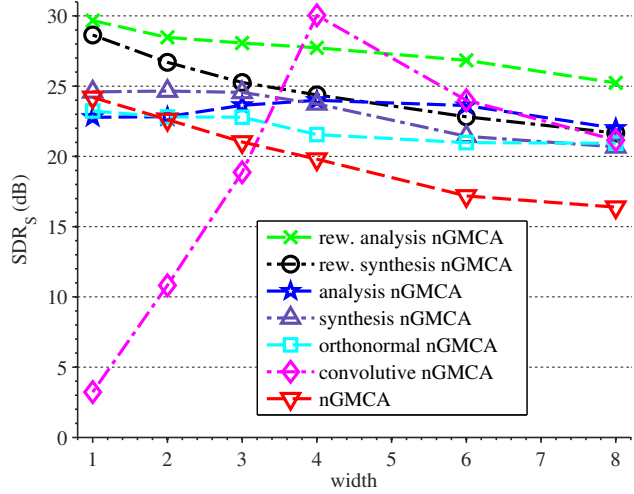}
\caption{Reconstruction SDR (SDR$_S$) with respect to the width of the peaks (simulated NMR spectra, SNR$_Y=20$dB,  $r=12$, average of $36$ simulations)}
\label{fig:width_nmr_rew}
\end{figure}

\subsubsection{Reweighting scheme on 1D spectra}

In this paragraph, the proposed reweighting scheme is applied to the simulated NMR spectra introduced in Section~\S\ref{sec:settings_eval}. Figures \ref{fig:noise_nmr_rew} and \ref{fig:width_nmr_rew} are therefore updates of \ref{fig:noise_nmr} and \ref{fig:width_nmr} with the addition of the reweighted analysis and synthesis formulations in the redundant wavelet domain. As expected, the reweighting procedure yields clear improvements in the analysis case with a gain of about 4 to 5 dB both in Figure \ref{fig:noise_nmr_rew} and \ref{fig:width_nmr_rew}. For the sake of consistency the exact same reweighting procedure is used for the synthesis formulation while it may not be as well suited in this case. Indeed, the synthesis formulation is much less constrained and may therefore benefit from harsher reweighting schemes. In \ref{fig:noise_nmr_rew} in particular, it can be seen that the procedure provides a slight improvement for this formulation under low noise conditions, but can hamper the performance in the high noise regime. In this case, the reweighting scheme may not sufficiently penalize small wavelet coefficients which are very likely to be mainly noise.

\begin{figure}[!t]
\centering
\includegraphics[width=3.35in]{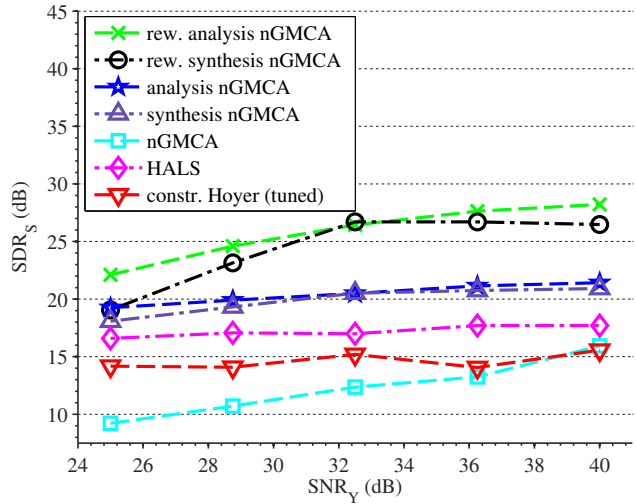}
\caption{Reconstruction SDR (SDR$_S$) with respect to the noise level in dB (simulated image mixtures, $r=4$, $m=16$, average of $24$ simulations)}
\label{fig:noise_images_rew}
\end{figure}

\subsubsection{Reweighting scheme on 2D natural images}
In this paragraph, numerical experiments are carried out to compare the analysis and synthesis sparse regularizations and their reweighted version in the context of imaging. In this setting, it is standard not to penalize the wavelet coarse scale which is usually not sparse since it encodes the very large scale of the sources. As a consequence, the coefficients of $\bs{\Lambda}$ which are related to the coarse scale are set to 0. In the same vein, the coarse scale does not generally contain discriminative information. It is therefore customary not to account for the corresponding wavelet coefficients to estimate the mixing matrix $\bA$. The wavelets used in these experiments are Daubechies-4 with 3 scales.\medskip

\begin{figure}[!t]
	\hspace{0.01\linewidth}
	\begin{subfigure}[b]{0.475\linewidth}
		\includegraphics[width=\linewidth]{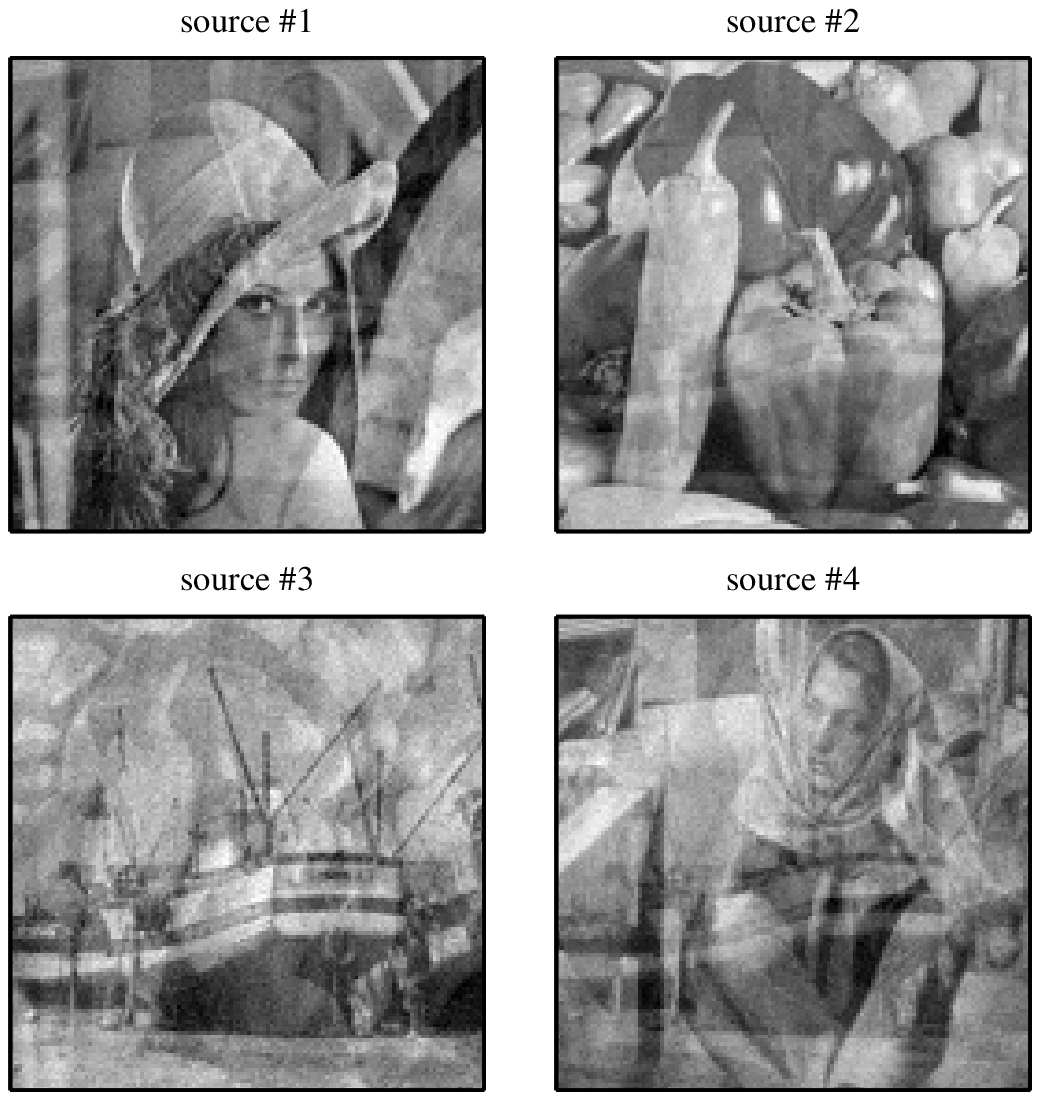}
	\caption{HALS\\(evaluating at SDR$_S = 14.79$dB).}
		\label{fig:reconstruction_2D_hals}
    \end{subfigure}%
    \hspace{0.03\linewidth}
	\begin{subfigure}[b]{0.475\linewidth}
		\includegraphics[width=\linewidth]{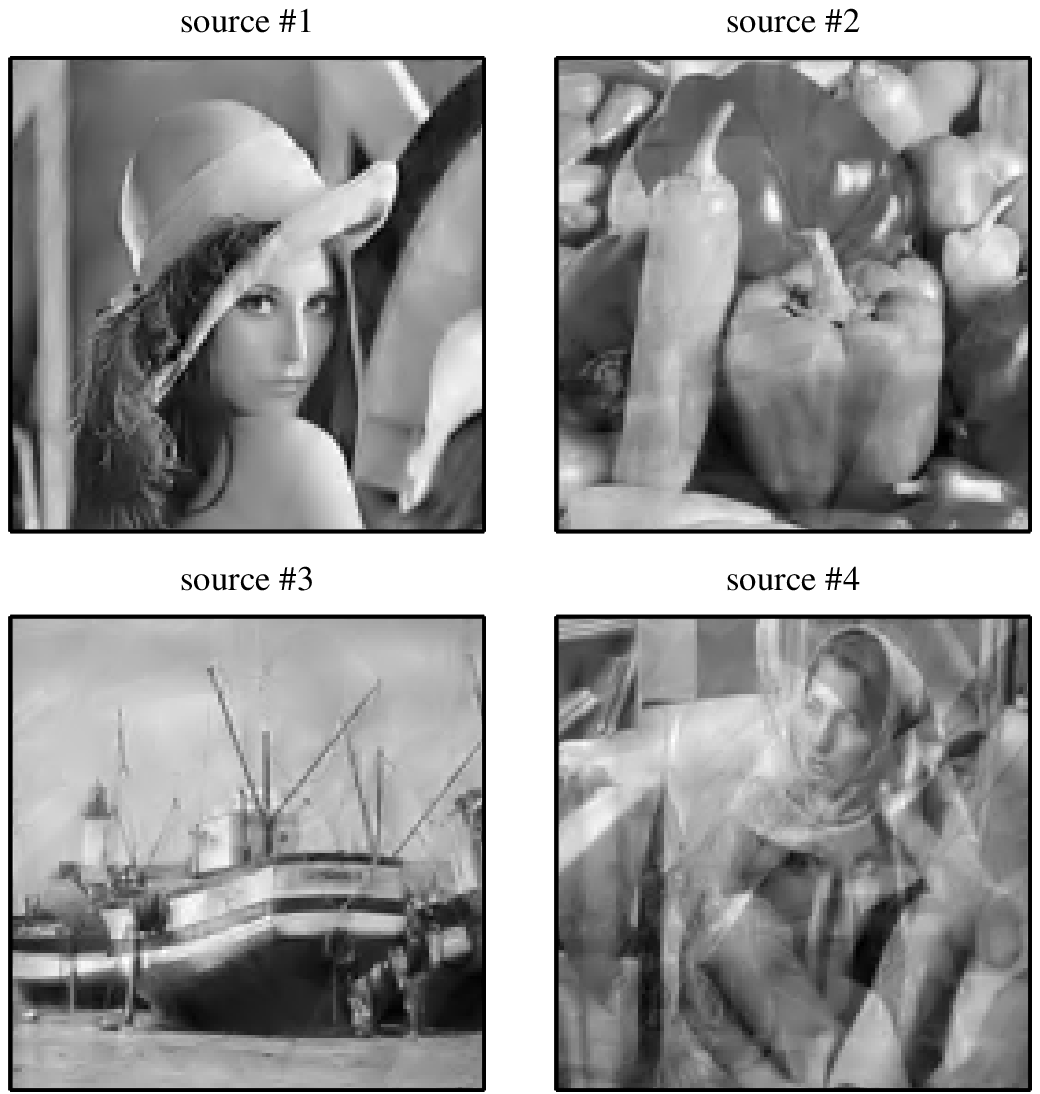}
		\caption{regular analysis nGMCA\\(evaluating at SDR$_S = 18.64$dB).}
		\label{fig:reconstruction_2D_ana}
	\end{subfigure}
	
	\hspace{0.2625\linewidth}
	\begin{subfigure}[b]{0.475\linewidth}
		\includegraphics[width=\linewidth]{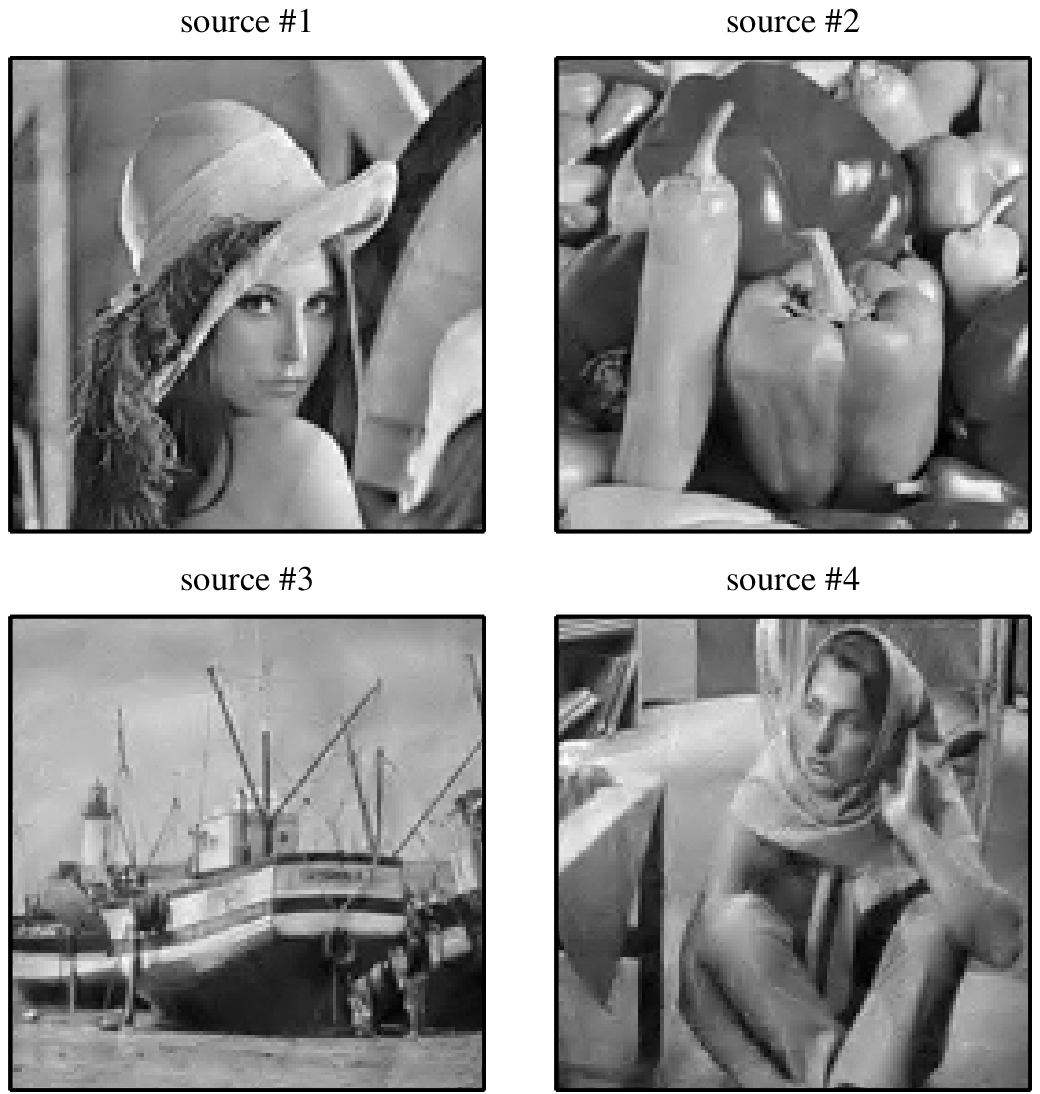}
		\caption{reweighted analysis nGMCA\\(evaluating at SDR$_S = 21.07$dB)}
		\label{fig:reconstruction_2D_ana_rew}
	\end{subfigure}	
	
	\caption{Examples of reconstruction ($16$ observations, SNR$_Y=30$dB).\vspace{-0.3cm}}
	\label{fig:reconstruction_2D}
\end{figure}

With the exception of the sources, the simulations are similar to the description in Section~\S\ref{sec:settings_eval}. Figure \ref{fig:noise_images_rew} provides the mean SDR of the reconstructions from the noisy mixtures for analysis and synthesis redundant wavelets with and without reweighted-$\ell_1$, as well as regular nGMCA and HALS. In this case, regular nGMCA gives poor results since the sources to be retrieved are clearly not very sparse in the direct domain. For this same reason, HALS is tested here with its sparse regularization. It performs better than nGMCA but is not able to denoise the sources and presents rather strong interferences, as displayed in the example of reconstruction provided in figure \ref{fig:reconstruction_2D_hals}.\\ 
In figure \ref{fig:reconstruction_2D_ana}, the solution provided by \ana_nGMCA without reweighting shows very limited noise contamination but exhibits strong interferences. This is particularly true in source \#4. Lastly, figure \ref{fig:reconstruction_2D_ana_rew} provides the reconstruction with the same algorithm but with the extra reweighting scheme. The proposed reweighting procedure yields much fainter interferences and still low noise.\medskip

This experiment corroborates our findings on much more complex data. The analysis formulation turns to be more robust than any other sparsity-enforcing regularization in the case of natural images which are mildly sparse in the wavelet domain. The use of the reweighting scheme, whether it is in the analysis or synthesis formulation, leads to a significant improvement of the separation performance of the nGMCA algorithm, as shown in figure \ref{fig:noise_images_rew}.

\bigskip
\section{Software}
\label{sec:software}

Following the philosophy of reproducible research \cite{Buckheit_95_WaveLabandReproducible}, the algorithms introduced in this article will be available at \url{http://www.cosmostat.org/GMCALab}. 

\bigskip
\section{Conclusion}
In this article, we have explored extensions of sparse NMF to cope with the blind separation of non-negative sources while promoting their sparsity in a transformed domain. To that end, we proposed a novel ensemble of algorithms, based on nGMCA, which can enforce sparsity in any sparsifying signal representation either in the synthesis or analysis formulation. Enforcing sparsity in some transformed domain as well as non-negativity in the direct domain is tackled by the use of recently introduced proximal calculus methods in convex optimization. To the best of our knowledge, this introduces the first application of the $\ell_1$ analysis regularization in the scope of blind source separation. Numerical experiments are carried out which show that the proposed algorithm provides enhanced separation performance for numerous types of sources which are neither sparse in the direct domain nor correctly modeled by smooth NMF methods. Furthermore, the analysis-based nGMCA algorithm is shown to outperform most sparsity-enforcing methods proposed so far in sparse NMF. Finally, we introduce a sparse reweighting procedure, either in the analysis and synthesis formulations, to further tackle the separation of complex data arising in imaging, which are generally mildly sparse. Numerical experiments have been conducted on natural images which show that the proposed sparse reweighting scheme yields clear improvements, especially in the analysis formulation.

\bigskip
\appendix

\newpage

\section{Norm-constrained Non-negative Proximal}
\label{app:norm_constrained_prox}

The aim is to find $\hat{y}$ such that:
\begin{align}
\hat{y}=\text{prox}_{.~\ge 0,~\|.\|_2^2\le 1}(x). \label{eq:normProx}
\end{align}
It is the well-defined proximal operator of a proper convex lower semi-continuous function as the characteristic function of a non-empty intersection of two convex sets. The Lagrangian of the corresponding optimization problem is given by:
\begin{equation}
\mathcal{L}(y,u,\lambda) = \frac{1}{2}\|y-x\|_2^2-\langle u,y \rangle + \frac{\lambda}{2}\left( \|y\|_2^2-1 \right).
\end{equation}

At optimum $(\hat{y}, \hat{u}, \hat{\lambda})$, the Karush-Kuhn-Tucker (KKT) conditions yield:
\begin{itemize}
\item $\hat{y} \ge 0$, $\|\hat{y}\|_2^2\le 1$,
\item $\hat{u}\ge 0$, $\hat{\lambda}\ge 0$,
\item $\hat{y}- x - \hat{u} +\hat{\lambda} \hat{y} = 0$,
\item $\hat{u}\odot \hat{y} = 0$, $\hat{\lambda} \left( \|\hat{y}\|_2^2-1 \right) = 0$.
\end{itemize}

Gathering all this together, we obtain:
\begin{equation}
\text{prox}_{.~\ge 0,~\|.\|_2^2\le 1}(x)=\frac{[x]_+}{\text{max}\left(\|~[x]_+\|_2,~1 \right)}.
\end{equation}
Proximal operator \eqref{eq:normProx} consequently boils down to a projection on the non-negativity constraint, followed by a projection on the $\ell_2$-unit-ball.

\bigskip
\section{Synthesis Proximal Operator}
\label{app:synProx}

The aim is to compute:
\begin{align}
\hat{y}_w=\text{prox}_{\bW^T\cdot\ge 0}(x_w), \label{eq:synProx}
\end{align}
for a column vector $x$ with $\bW$ an operator such that $\bW^T \bW=\bI$. This proximal operator is well-defined since $i_{\bW^T\cdot\ge 0}$ is a proper convex and lower semi-continuous function. Since $\underset{u \le 0}{\text{max}}~\langle u, \bW^T y_w \rangle =i_{\cdot\ge 0}(\bW^T y_w)$, computing this proximal operator is equivalent to solving:

\begin{equation}
\underset{y_w}{\text{argmin}}~\underset{u \le 0}{\text{max}}~
 \frac{1}{2}\|y_w - x_w\|_2^2+\langle u, \bW^T y_w \rangle.
\end{equation}

Considering the Von Neumann min-max theorem, the min and max operators can be inverted. Finding the  optimal $\hat{y}_w$ which minimizes the cost function for a given $u$ yields the relationship: $\hat{y}_w = x_w - \bW u$. Replacing $y_w$ by its optimal value and rearranging the terms leads to the following optimization problem:

\begin{equation}
\label{app:syn_dual_problem}
\underset{u \le 0}{\text{argmin}}~
 \frac{1}{2}\| \bW u - x_w \|_2^2.
\end{equation}
Since $\bW^T\bW = \bs{I}$, this is equivalent to the minimization of:

\begin{equation}
\underset{u \le 0}{\text{argmin}}~
 \frac{1}{2}\| u - \bW^Tx_w \|_2^2,
\end{equation}
which is straightforwardly the projection of $\bW^T x_w$ on the set of matrices with non-positive coefficients: $ \hat{u} =  -\big [-\bW^T x_w\big ]_+$.\\
Consequently, gathering everything together provides an analytic solution for the searched proximal operator:
\begin{equation}
\text{prox}_{i_{\cdot\ge 0}(\bW^T.)}(x_w) = x_w + \bW\big [-\bW^T x_w\big ]_+.
\end{equation}

\section{Synthesis and Non-negative Update Step}
\label{app:syn_update}

\textbf{Algorithm \ref{alg:synthesis_update}} below implements the synthesis and non-negative update of $\bS$:
\begin{equation}
\underset{ \bS_w \bW \ge \b0}{\text{argmin}}~\frac{1}{2}\|\bY-\bA\bS_w\bW \|_2^2  + \| \bs{\Lambda}^{(k)} \odot  \bS_w\|_1.
\end{equation}
It converges if $\gamma < \frac{2}{L}$ and $\mu \in \left]0,\text{min}\left(\frac{3}{2},\frac{1+2/(L\gamma)}{2}\right)\right[$  where $L = \|\bA^T\bA\|_{s,2}$, and uses the Generalized Forward-Backward algorithm \cite{Raguet_13_GeneralizedForwardBackward}.

\begin{algorithm}[h]
\caption{: Synthesis update}
\label{alg:synthesis_update}
\begin{algorithmic}[1]
\State \textbf{initialize} $\bS_w^{(0)}$, $\bs{U}_w^{(0)}$, $\bs{V}_w^{(0)}$ $\gamma>0$, $\omega_u,\omega_v  \in \left]0,1\right[$ such that $\omega_u + \omega_v = 1$, $k=0$.
\While{(not converged)}
\State $\bs{G}_w = \bA^T\left(\bA\bS_w\ik\bW-\bY\right)\bW^T$
\Comment gradient computation
\State $\bs{U}_w\ikp = \bs{U}_w\ik - \mu \bS_w\ik + \mu \text{Soft}_{\frac{\gamma}{\omega_u}\bs{\Lambda}}
\left(2\bS_w\ik-\bs{U}_w\ik-\gamma \bs{G}_w\right)$
\State $\bs{V}_w\ikp = \bs{V}_w\ik - \mu \bS_w\ik + \mu \text{prox}_{i_{\cdot\ge 0}(\bW^T.)}
\left(2\bS_w\ik-\bs{V}_w\ik-\gamma \bs{G}_w\right)$
\State $\bS_w\ikp = \omega_u \bs{U}_w\ikp + \omega_v \bs{V}_w\ikp$
\State $k = k + 1$
\EndWhile
\State \textbf{return} $\bS_w^{(k)}$
\end{algorithmic}
\end{algorithm}

\bigskip
\section{Analysis proximal operator}
\label{app:anaProx}

The aim is to find $\hat{y}$ such that:
\begin{align}
\hat{y}=\text{prox}_{\|\lambda \odot \bW \cdot\|_1}(x), \label{eq:anaProx}
\end{align}
with $x$ a column vector and $\bW$ a matrix transform on $x$. This proximal operator is well-defined since $\|\lambda \odot \bW \cdot\|_1$ is a proper convex and lower semi-continuous function. Using the fact that:
\begin{equation}
\underset{|u_{w|i}| \le \lambda_i}{\text{max}}  \langle u, \bW y \rangle =\|\lambda \odot \bW y\|_1, 
\end{equation}
and following the same steps as in \ref{app:synProx} up to equation \eqref{app:syn_dual_problem}, we obtain the relationship $\hat{y} = x - \bW^T u_w$, and that the computation of \ref{eq:anaProx} is equivalent to solving the following problem:
\begin{equation}
\label{app:ana_dual_problem}
\underset{|u_{w|i}| \le \lambda_{i}}{\text{argmin}}~
 \frac{1}{2}\| \bW^T u_w - x\|_2^2,
\end{equation}

However, in opposition to the synthesis case, $\bW\bW^T\ne \bI$ and problem \eqref{app:ana_dual_problem} does not have any analytical solution. It can however be computed using the forward-backward algorithm, with gradient $u_w \mapsto \bW(\bW^T u_w - x)$ and proximal $P^\infty_{\lambda}:u_w \mapsto u_w-\text{Soft}_{\lambda}(u_w)$ (projection onto the $\ell_\infty$-ball).\\
Finally, we obtain:
\begin{equation}
\text{prox}_{\| \lambda \odot \bW \cdot\|_1}(x)=x- \bW^T\underset{|u_{w|i}| \le \lambda_{i}}{\text{argmin}}~\frac{1}{2}\| \bW^Tu_w - x \|_2^2.
\end{equation}

\bigskip
\section{Analysis and Non-negative Update Step}
\label{app:analysis_chambolle}

The problem:
\begin{align}
\underset{\bS \ge \b0}{\text{min}}~
 \frac{1}{2}\|\bY - \bA \bS\|_2^2+\|\bs{\Lambda} \odot  (\bS \bW^T)\|_1, \label{eq:S_analysis}
\end{align}
can be reformulated under the settings of \cite{Chambolle_10_firstorderprimal}:
\begin{align}
\underset{\bS}{\text{min}} \underset{| \bs{U}_{w|i,j} | \le \bs{\Lambda}_{i,j}}{\text{max}} \underset{\bs{V} \le \b0}{\text{max}}~\frac{1}{2}\|\bY - \bA \bS\|_2^2+ \langle \bS \bW^T,\bs{U}_w \rangle  +  \langle \bS, \bs{V} \rangle .   \label{eq:S_analysis_CP}
\end{align}
Hence, it can be written
\begin{align}
\underset{\bS}{\text{min}}~\underset{\bs{U}_w, \bs{V}}{\text{max}}~G(\bS)+ F_1^*( \bs{U}_w)+ F_2^*(\bs{V}) + \langle \bS \bW^T,\bs{U}_w \rangle  +  \langle \bS, \bs{V} \rangle , \label{eq:S_analysis_CP2}
\end{align}
with:
\begin{align}
&G(\bS) =\frac{1}{2}\|\bY - \bA \bS \|_2^2. \notag \\
&F_1^*( \bs{U}_w) = i_{| \bs{U}_{w|i,j} | \le {\bs{\Lambda}}_{i,j}}(\bs{U}_w). \notag \\
&F_2^*(\bs{V} ) = i_{\bs{V} \le \b0}(\bs{V}). \notag
\end{align}
Notice that, the dual variable has been split into two for the sake of simplicity, but they could be easily gathered into a unique variable.\medskip

The algorithm then requires the knowledge of the proximal operator of $G$ which is straightforwardly given by:
\begin{align}
\text{prox}_{\tau G}(\bS) &= \underset{\bs{R}}{\text{argmin}}~\frac{\tau}{2}\|\bY - \bA \bs{R} \|_2^2+ \frac{\text{1}}{\text{2}}\|\bs{R} - \bS \|_2^2,\notag \\
& = (\bs{I}+\tau \bA^T\bA)^{-1}(\bS+\tau \bA^T\bY), \label{eq:prox_L2}
\end{align}
and the proximal operators of $\sigma F_1$ and $\sigma F_2$ which are respectively projections on the $\ell_\infty$-ball $P^\infty_\Lambda$ (proximal \#\ref{prox:Linf}) and on the non-positive constraints (proximal \#\ref{prox:pos}).\medskip

\textbf{Algorithm \ref{alg:analysis_update}} below therefore converges to a solution of problem \ref{eq:S_analysis} if $\tau\sigma L^2<1$, where $L = \sqrt{1 + \|\bW\|^2_{s,2}}$. If $\bW$ is a tight frame with $\bW^T\bW = \bs{I}$, which is the case in our experiments, $L=\sqrt{2}$.
 
\begin{algorithm}[h]
\caption{: Analysis update}
\label{alg:analysis_update}
\begin{algorithmic}[1]
\State \textbf{initialize} $\bS^{(0)}$, $\bs{U}_w^{(0)}$, $\bs{V}^{(0)}$, $\bar{\bS}^{(0)}=\bS^{(0)}$,  $\tau >0,~\sigma >0$, $k=0$.
\While{(not converged)}
\State $\bs{U}_w\ikp = P^\infty_\bs{\Lambda}\left(\bs{U}_w\ik+\sigma\bar{\bS}\ik\bW^T\right)$
\State $\bs{V}\ikp = -\left[-\bs{V}\ik-\sigma\bar{\bS}\ik\right]_+$
\State $\bS\ikp = (\bs{I}+\tau \bA^T\bA)^{-1}\left(\bS \ik -\tau (\bs{U}_w\bW+\bs{V}-\bA^T\bY)\right)$
\State $\bar{\bS}\ikp = 2\bS\ikp - \bS\ik$
\State $k = k + 1$
\EndWhile
\State \textbf{return} $\bS^{(k)}$
\end{algorithmic}
\end{algorithm}

\newpage


\end{document}